%% file: paper.tex
\documentclass{article}
\usepackage{silence}
\usepackage[preprint]{neurips_2026}

\usepackage[utf8]{inputenc} 
\usepackage[T1]{fontenc}    
\usepackage{booktabs}       
\usepackage{xcolor}
\usepackage[title]{appendix}
\usepackage{microtype}
\usepackage{graphicx}
\usepackage{subcaption}
\usepackage{booktabs}
\usepackage{placeins}
\usepackage{pdflscape}
\usepackage{hyperref}
\usepackage{url}
\usepackage{booktabs}
\usepackage{multirow}

\newcommand{\ours}{\textsc{ac-gpt}}
\newcommand{\sigmagpt}{\textsc{$\sigma$-gpt}}
\newcommand{\gpt}{\textsc{gpt}}
\newcommand{\mlm}{\textsc{mlm}}
\newcommand{\diffusion}{\textsc{diffusion}}
\newcommand{\ie}{\textit{i.e.}}
\newcommand{\eg}{\textit{e.g.}}

\usepackage{array}
\usepackage{pifont}
\newcommand{\cmark}{\ding{51}}
\newcommand{\xmark}{\ding{55}}
\newcolumntype{C}{>{\centering\arraybackslash}p{2.3cm}}

\usepackage{enumitem}
\setlist[itemize]{itemsep=3pt, topsep=0pt, parsep=0pt, partopsep=0pt}
\setlist[enumerate]{itemsep=3pt, topsep=0pt, parsep=0pt, partopsep=0pt}

\usepackage{amsmath}
\usepackage{amsfonts}
\usepackage{amssymb}
\usepackage{mathtools}
\usepackage{mathrsfs}
\usepackage{nicefrac}
\usepackage{amsthm}
\usepackage{array, makecell}
\usepackage{listings}

\usepackage[nameinlink,capitalize,noabbrev]{cleveref}

\theoremstyle{plain}

\theoremstyle{definition}

\theoremstyle{remark}

\usepackage[most]{tcolorbox}
\usepackage{soul}
\sethlcolor{gray!25}
\newtcolorbox[auto counter]{mybox}[2][]
{
  center,
  breakable,
  width = 1.09\linewidth,
  colframe = gray!25,
  colback  = gray!10,
  coltitle = black,  
  title    = \textbf{Box \thetcbcounter. #2},
  #1,
}

\usepackage[textsize=tiny]{todonotes}

\title{Simplifying the Modeling of Arbitrary Conditionals\\ in Natural Language}

\author{
  Yinhan Lu${}^{*,1,2,}$,
  Eric Elmoznino${}^{*,1,3}$,\\
  \textbf{Léo Gagnon${}^{1,3}$,}
  \textbf{Sarthak Mittal${}^{1,3}$,}
  \textbf{Tejas Kasetty${}^{1,3}$,}
  \textbf{Guillaume Lajoie${}^{1,3}$} \\\\
  ${}^1$Mila --- Quebec AI Institute,
  ${}^2$McGill University,
  ${}^3$Université de Montréal
  \vphantom{
    \thanks{Equal contribution. Correspondence to: \texttt{\{yinhan.lu,eric.elmoznino,guillaume.lajoie\}@mila.quebec}.}
  }
}

\begin{document}

\maketitle

\begin{abstract}
  Causal Transformers model sequences through an autoregressive factorization of the joint distribution, which enables efficient left-to-right decoding and conditional likelihood computation. However, they cannot tractably sample from or evaluate arbitrary conditionals --- \eg, a block of text conditioned on past \textit{and future} tokens. Recent work aims to solve this problem through novel architectures, but they often lead to sub-optimal modeling of such conditionals and degraded generations. We propose \textbf{A}rbitrary \textbf{C}onditionals \textbf{GPT} (\ours{}) which introduces a simple modification to standard causal Transformers to enable evaluating and sampling from arbitrary conditionals --- including past, future, and mixed contexts --- within a single forward pass. Unlike prior approaches, our method preserves the standard left-to-right ordering and next-token prediction objective essential for both strong performance and efficient training on natural language. Crucially, this compatibility allows existing LLMs to be fine-tuned for arbitrary conditioning. Our empirical results indicate that our method outperforms baselines on modeling arbitrary conditionals, without degrading standard left-to-right performance.
\end{abstract}

\section{Introduction}

State-of-the-art large language models (LLMs) work by predicting the next token in a sequence given past tokens. Causal Transformers --- the most successful and widely-used architecture for language modeling --- implement this efficiently by evaluating the joint likelihood of a sequence under a left-to-right factorization in a single forward pass \citep{radford2019language}. As a result of their left-to-right structure, these models are also capable of evaluating and sampling from temporally ordered conditional probabilities, \ie, the likelihood of future tokens conditioned on past ones, $p(\mathbf{x}_{> t} \mid \mathbf{x}_{\le t})$.
However, evaluating \textit{arbitrary conditionals} $p(\mathbf{x}_e \mid \mathbf{x}_c)$ defined by an arbitrary set of observed tokens $\mathbf{x}_c$, would amount to an intractable integral \citep{hu2024amortizing}.
The evaluation of arbitrary conditionals can serve a number of downstream applications such as infilling, text-editing (\eg, replace a word, sentence, or paragraph in the context of a larger body of text), reasoning towards a known conclusion, and other tasks that require conditioning on more than just past tokens. Recent work has introduced methods to facilitate these tasks in sequence modeling, but current approaches rely on bespoke neural architectures, and can be cumbersome to leverage at scale \citep{yang2019xlnet,pannatier2024sigma}.

In this work, we introduce Arbitrary Conditionals GPT (\ours{}) --- a simple modification to the standard causal Transformer architecture that augments it with the ability to evaluate and sample from arbitrary conditionals, including conditioning on the past, future, and mixtures of both. In contrast to other approaches \citep[\eg,][]{devlin2018bert,pannatier2024sigma}, our method preserves the left-to-right structure of the sequence and can evaluate any conditional in a single forward pass, both of which are necessary for strong performance on natural language. We achieve this by preserving the autoregressive next-token-prediction objective through causal attention, while modeling arbitrary conditionals by augmenting the context with conditioned tokens, regardless of where they appear in the sequence.

Through extensive evaluation, we show that \ours{} consistently yields superior performance to existing baselines on diverse conditional likelihood evaluations and sampling tasks, without degrading the performance on left-to-right modeling that standard causal Transformers provide. Furthermore, given the structural similarity of \ours{} to most current language models, we show that we can easily fine-tune LLMs to augment them with the capability of modeling arbitrary conditionals.

\section{Background and Related Work}
\label{sec:background}

In this section, we review existing approaches to the problem of evaluating and sampling from arbitrary conditionals in natural language, highlighting key pitfalls addressed by our method. These comparisons are summarized in \cref{tab:comparisons}.

\begin{table*}[t]
\centering
\small
\begin{tabular}{l @{\hspace{1pt}} C @{\hspace{1pt}} C @{\hspace{1pt}} C @{\hspace{1pt}} C @{\hspace{1pt}} C @{}}
\toprule
Method & Evaluate any conditional & Sample any conditional & Left-to-right decomposition & Single-pass evaluation & Fine-tune GPT-style LLM \\
\midrule
    \textbf{\ours{} (ours)} & \cmark & \cmark & \cmark & \cmark & \cmark \\
    \gpt{} & \xmark & \xmark & \cmark & \cmark & --- \\
    \sigmagpt{} & \cmark & \cmark & \xmark & \cmark & \xmark \\
    \mlm{} & \cmark & \cmark & \xmark & \xmark & \xmark \\
    \diffusion{} & \xmark & \cmark & \xmark & \cmark & \xmark \\
\bottomrule
\end{tabular}
\caption{\textbf{Comparison of our method, \ours{}, to other approaches.} \gpt{} refers to a standard left-to-right autoregressive Transformer \citep{radford2019language}. \sigmagpt{} is a representative any-order autoregressive causal Transformer \citep{pannatier2024sigma}. \mlm{} refers to a Transformer with unrestricted bidirectional attention that unmasks multiple tokens in parallel conditioned on observed ones \citep{devlin2018bert}. \diffusion{} refers to discrete diffusion methods \citep[\eg,][]{sahoo2024simple,austin2021structured,shi2024simplified}.}
\label{tab:comparisons}
\end{table*}

\paragraph{Causal Transformers.}

Standard causal Transformers, such as the GPT family of architectures \citep{radford2019language}, model the joint distribution of a sequence $\mathbf{x}$ autoregressively via the chain rule, factorized in a fixed left-to-right order: $p(\mathbf{x}) = \prod_{t} p(x_t \mid \mathbf{x}_{<t})$, where each token's final hidden representation is used to predict the next token in the sequence. This is implemented efficiently using causal masking and teacher forcing, allowing the model to evaluate the likelihood of an entire sequence in a single forward pass during training. Due to their alignment with the natural temporal structure of language, these models have achieved immense success in scaling to large datasets, powering modern prompting and chat-based applications \citep{achiam2023gpt}. However, because they are structurally bound to a left-to-right decomposition, they are fundamentally unable to tractably evaluate conditional probabilities where the target tokens precede the conditioning tokens (\eg, $p(\mathbf{x}_{\text{middle}} \mid \mathbf{x}_{\text{future}})$) --- a requirement for tasks such as infilling or constrained text editing. \citet{bavarian2022efficient} trains causal Transformers to do infilling through a data augmentation process that prompts them with the start and end of a sequence, but cannot model other kinds of conditionals.

\paragraph{Any-order autoregressive Transformers.}

Several works have attempted to address arbitrary conditioning by modifying the causal Transformer to support \textit{any} autoregressive ordering, rather than a fixed left-to-right order. Prominent examples like XLNet \citep{yang2019xlnet} and \sigmagpt{} \citep{pannatier2024sigma} achieve this by (a) randomly sampling permutation orders during training and (b) modifying the Transformer architecture so that each token is told not only what context it may attend to, but also which original sequence position it is supposed to predict under a non-standard autoregressive order. For example, \sigmagpt{} uses a double positional encoding scheme --- one for the token's content position and one for its decoding position. 

We argue that these approaches have several downsides. First, the additional target-position machinery makes the prediction problem substantially harder: the model must learn token representations that support prediction at many possible target locations, rather specializing on the next token which is thought to be an especially good choice for language modeling \citep{papadopoulos2024arrows}. Second, notice that on top of supporting arbitrary conditionals in terms of what tokens are observed, this class of methods supports arbitrary \emph{orderings} within the observed and predicted tokens, which wastes model capacity.

For instance, consider the sequence $(x_1, x_2, x_3, x_4, x_5)$ and the conditional $p(x_1, x_2 \mid x_3, x_4, x_5)$. An autoregressive arbitrary-order model $p_\theta$ would attempt to model this query under all possible \emph{orderings}, including $p(x_2 \mid x_3, x_4, x_5)p(x_1 \mid x_3, x_4, x_5, x_2)$, $p(x_1 \mid x_3, x_4, x_5)p(x_2 \mid x_3, x_4, x_5, x_1)$, $p(x_2 \mid x_5, x_4, x_3)p(x_1 \mid x_5, x_4, x_3, x_2)$ and so on, even though these all represent the same conditional probability query. Crucially, language data has a natural left-to-right ordering under which it is more easily modeled \citep{papadopoulos2024arrows}, so attempting to model all possible orderings effectively wastes model capacity on worse orderings. Indeed, \citet{pannatier2024sigma} observed that without a curriculum favoring left-to-right orderings, any-order models often fail to learn effectively on complex datasets. In contrast, our method decomposes every conditional probability query into a single left-to-right factorization, spending all its capacity on good ordering for natural language.

\paragraph{Masked language models.}

Masked Language Models (MLMs) like BERT \citep{devlin2018bert} utilize bidirectional attention and are trained to predict masked tokens given unmasked context. While effective for representation learning, they are not naturally generative. During training, MLMs maximize the likelihood of independent marginals over masked tokens, rather than their joint probability. Consequently, they cannot evaluate the joint likelihood of a multi-token segment in a single pass. Generating or evaluating text with MLMs requires iterative procedures --- such as Gibbs sampling or multi-step unmasking --- which are computationally expensive \citep{wang2019bert}. Furthermore, like any-order autoregressive models, MLMs implicitly attempt to model dependencies in all directions, incurring similar capacity constraints that limit their effectiveness on natural language data compared to strictly causal baselines.

\begin{figure*}[ht]
    \centering
    \includegraphics[width=\linewidth]{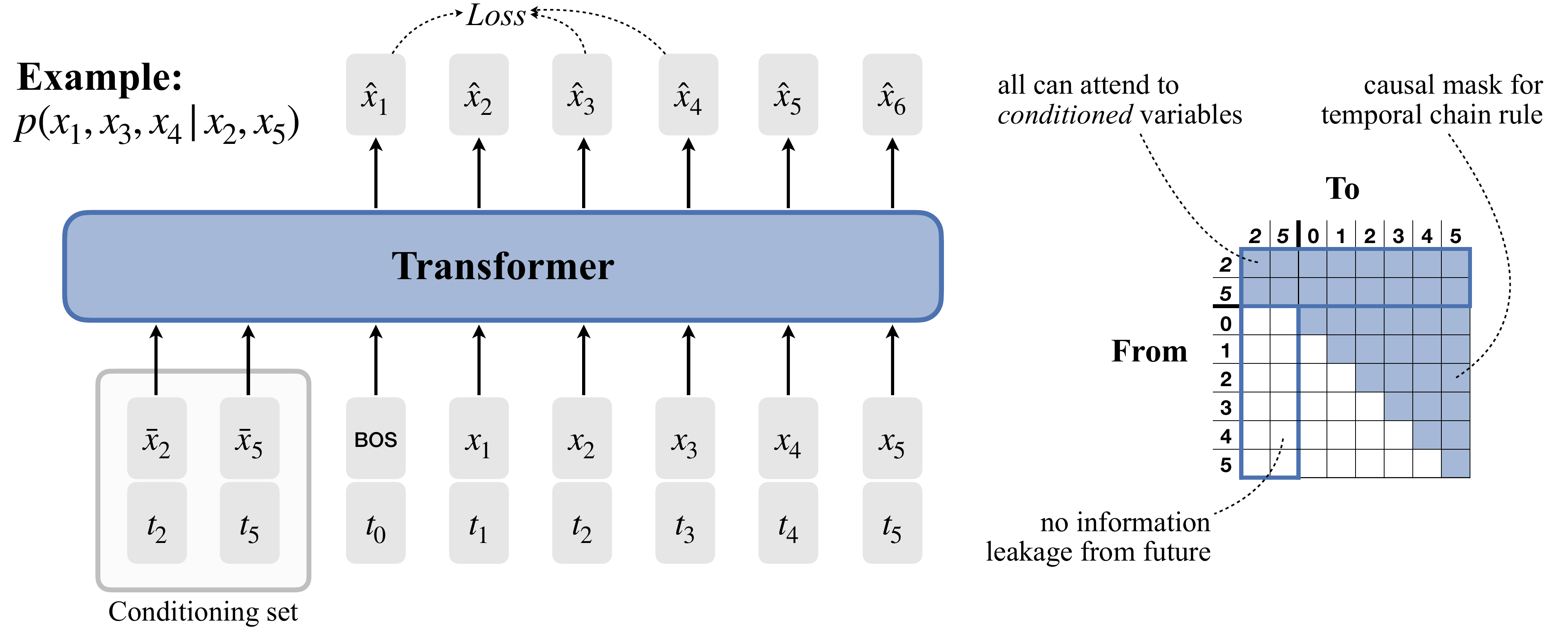}
    \caption{\textbf{Illustration of our method, \ours{}.} We train a standard causal Transformer architecture, but augment the sequences with a set of known tokens representing the \textit{conditioning set} which all other tokens can attend to. Each token's final hidden representation is used to predict the next token in the sequence, and the loss for tokens in the \textit{evaluation set} are used to update the model. \texttt{BOS} is the \texttt{<BEGINNING OF SEQUENCE>} token, and $\bar{x}_i$ represents a copy of a token with identical embedding and positional encoding. We train over a joint distribution of conditioning/evaluation sets, each representing a conditional probability query.}
    \label{fig:model}
\end{figure*}

\paragraph{Discrete diffusion models.}

Discrete diffusion models have recently emerged as a competitive alternative for non-autoregressive text generation. Approaches such as D3PM \citep{austin2021structured} and masked diffusion language models (MDLMs) \citep{sahoo2024simple, shi2024simplified} generate text by iteratively denoising or unmasking a sequence. While these models can sample from arbitrary conditionals by fixing observed tokens and denoising the rest, they suffer from two key limitations relevant to our goals. First, they can sample from but cannot tractably \textit{evaluate} the likelihood of a sequence or conditional, which makes them unsuitable for applications requiring precise density estimation, such as rejection sampling or planning. Second, they typically lack the inductive bias of left-to-right ordering, which empirically harms their performance \citep{xue2025any}. Even state-of-the-art MDLMs generally underperform standard autoregressive Transformers in terms of perplexity \citep{sahoo2024simple}, likely due to the difficulty of optimizing a variational lower bound of the log likelihood without the structural guidance of the left-to-right decomposition inherent to natural language.

Crucially, all the aforementioned approaches --- any-order Transformers, MLMs, and diffusion models --- require fundamental architectural deviations from the standard causal Transformer. This makes it infeasible to adapt pre-trained LLMs to these paradigms without training from scratch. For example, \sigmagpt{} requires a second positional embedding table for decoding positions that does not exist in pretrained LLMs, and the use of these decoding positional embeddings would need to be learned from scratch. In contrast, our method preserves the architecture of a causal Transformer, enabling fine-tuning of existing LLMs for arbitrary conditioning while still leveraging the left-to-right structure inherent to natural language.

\section{Modeling Arbitrary Conditionals}
\label{sec:methods}

We now describe our approach to modeling arbitrary conditionals, which we term \textbf{A}rbitrary \textbf{C}onditionals \textbf{GPT} (\ours{}) and illustrate in \cref{fig:model}.

\paragraph{Base model.}

\ours{} leverages a standard causal decoder-only Transformer architecture \citep{radford2019language}, and models arbitrary conditionals by augmenting the sequence with additional tokens. The main features of a causal Transformer are that (a) tokens can only attend to the past, (b) each token's hidden representation at the final layer is used to decode the following token in the sequence, (c) the model is trained using teacher forcing, and (d) positional information is encoded in the token embeddings. We use a GPT-2 architecture, but with rotary as opposed to relative positional encodings \citep[RoPE,][]{su2024roformer} due to their improved performance.

\paragraph{Conditional queries.}

To model a conditional query, we consider a sequence $\mathbf{x} = (x_1, ..., x_T)$ that is partitioned into two non-overlapping subsets: a \textit{conditioning set} $\mathbf{x}_c$ representing conditioning tokens and an \textit{evaluation set} $\mathbf{x}_e$ representing the tokens over which we wish to evaluate the joint distribution. Importantly, the sequence elements belonging to each partition can be arbitrarily set, and original sequence ordering information is conserved. Our goal is to model the resulting query $p(\mathbf{x}_e \mid \mathbf{x}_c)$ for any partitioning of $\mathbf{x}$ into $(\mathbf{x}_e, \mathbf{x}_c)$. At training time, the loss only includes the average prediction error for tokens in $\mathbf{x}_e$ --- predictions for conditioning tokens cannot be used because the model already has access to them.

\ours{} handles conditioning tokens $\mathbf{x}_c$ by creating \textit{copies} $\bar{\mathbf{x}}_c$ which are placed at the ``start'' of the sequence, in the sense that they can be attended to by all other tokens in the sequence. Though this new ordering is used by causal attention, the copies $\bar{\mathbf{x}}_c$ maintain the position encodings of their original location in the sequence (see \cref{fig:model}). This effectively augments the sequence with a prompt containing position-aware conditioning information. Because conditioning tokens are known, there is no need to maintain a causal attention mask within the conditioning set. Therefore, to maintain greater expressive power, we use unrestricted bidirectional attention between tokens in $\bar{\mathbf{x}}_c$. We also note that a similar copying mechanism is used in block diffusion for efficient training \citep{arriola2025block}.

It is important to clarify why \ours{} must include \textit{copies} $\bar{\mathbf{x}}_c$ of conditioning tokens that all others can attend to, rather than just allowing all tokens to attend to the original tokens $\mathbf{x}_c$ that are already present in the sequence. Effectively, these copies are necessary to prevent information leakage across multiple Transformer layers when each token makes a prediction for the next one in the sequence. This is a subtle point that is best illustrated using a concrete example. Consider a sequence $\mathbf{x} = (x_1, x_2, x_3, x_4)$ where $\mathbf{x}_e = \{x_1, x_2, x_4\}$ and $\mathbf{x}_c = \{x_3\}$. In order for $x_3$ to make a valid prediction $\hat{x}_4$ for its next token, it must be able to attend to all tokens before it. In addition, both $x_1$ and $x_2$ must be able to attend to the conditioning token $x_3$. Without our copying mechanism, then, information about $x_2$ would be transmitted to $x_3$ after one layer through standard causal attention, and would subsequently be transmitted to $x_1$ through $x_3$ in the following layer. This indirect information leakage from $x_2$ to $x_1$ violates the left-to-right chain rule decomposition of the sequence. For this reason, we create a copy $\bar{\mathbf{x}}_c$ of the conditioning set that the evaluation set $\mathbf{x}_e$ can attend to over multiple layers, \textit{without} leaking information from later tokens of $\mathbf{x}_e$ into earlier ones.

In summary, as shown in \cref{fig:model}, the complete augmented sequence that is passed to our model is $[\bar{\mathbf{x}}_c, \mathbf{x}]$ where:
\begin{itemize}
    \item $\bar{\mathbf{x}}_c$ is a copy of $\mathbf{x}_c$.
    \item Causal attention is used everywhere except for within $\bar{\mathbf{x}}_c$, which uses bidirectional attention.
    \item The training loss only includes the average prediction error on tokens in $\mathbf{x}_e$.
\end{itemize}

\paragraph{Conditioning set distribution.}

In order to model truly arbitrary conditional queries at inference time, we would need to train on a distribution of conditional queries with full coverage. However, modeling more varied queries requires more model capacity and training time \citep{shih2022training}. In reality, there is often structure in the kinds of conditional queries a model needs to evaluate, depending on the intended downstream tasks --- for instance, perhaps only a small subset of the sequence needs to be conditioned on, or perhaps conditioning tokens tend to be part of contiguous subsequences.

In this work, we sample our conditioning set $\mathbf{x}_c$ from a distribution that interpolates over fully arbitrary conditional queries and more structured ones (as illustrated in \cref{fig:arbitrary-conditioning} likely to subserve downstream tasks. Intuitively, our sampling procedure generates conditioning sets consisting of varying numbers of contiguous blocks of varying sizes. Concretely, for any given training sequence $\mathbf{x}$ of length $|\mathbf{x}|$, we sample a conditioning set $\mathbf{x}_c \subset \mathbf{x}$ as follows:
\begin{enumerate}[leftmargin=*]
    \item First, we sample the length of the conditioning set $|\mathbf{x}_c|$ within a range dependent on the sequence length. Specifically, $|\mathbf{x}_c| \sim \text{Unif}(r_{min} |\mathbf{x}|, r_{max} |\mathbf{x}|)$, where $0 \le r_{min} \le r_{max} \le 1$ are fixed hyperparameters.

    \item Next, we sample the number of contiguous conditioning blocks $B \sim \text{Unif}(b_{min},\; b_{max})$, with $1 \le b_{min} \le b_{max} \le |\mathbf{x}_c|$. Both $b_{min}$ and $b_{max}$ are either fixed hyperparameters or variables whose range depends on $|\mathbf{x}_c|$.

    \item Then, we sample the block sizes $\{s_i\}_{i=1}^B$ (such that $\sum_{i=1}^B s_i = |\mathbf{x}_c|$) by initializing $s_i = 1$ for all $i$ and distributing each of the remaining $|\mathbf{x}_c| - B$ tokens uniformly and independently into one of the B blocks. This procedure produces block sizes that are typically balanced in contrast to uniformly sampling block sizes subject to the same sum constraint, which often yields highly uneven partitions.
    \item Finally, we sample the locations of the $B$ blocks by drawing $B+1$ gap lengths (before the first block, between blocks, and after the last block). This is obtained by sampling $B$ values without replacement in the range $1$ to $B + |\mathbf{x}_e|$ --- the space between these values (and between the values and range boundaries) are the gap sizes.
\end{enumerate}

Under this sampling procedure, setting $(r_{min} = 0,\; r_{max} = 1,\; b_{min} = b_{max} = |\mathbf{x}_c|)$ results in full coverage over possible conditional queries with no inductive bias over which are more common. On the other hand, narrowing the range of conditioning set sizes and number of blocks results in less coverage over possible conditional queries and greater inductive bias for contiguous conditioning structure. In this way, our parameterization interpolates between arbitrary and structured conditional queries.

We note that the distribution from which the conditioning set is sampled is a design choice, and the optimal one depends on the downstream task. For instance, if the downstream task is infilling with a single contiguous evaluation set, the conditioning set could be constructed by sampling the beginning and end of the sequence, \ie, the prefix and suffix of the evaluation set.

\section{Experiments}

We now show empirical experiments that demonstrate \ours{}'s capabilities. In \cref{sec:results}, we evaluate \ours{} against existing baselines on natural language and demonstrate that it achieves superior performance on conditional queries while preserving standard left-to-right capability. In \cref{sec:fine-tuning}, we then show that \ours{}'s architectural compatibility uniquely enables direct fine-tuning of pretrained large language models, where it yields substantial perplexity gains on conditional queries at billion-parameter scale.

\subsection{Experimental Setup}
\label{sec:experimental_setup}

Our primary evaluation is on the Wikitext-103 natural language dataset \citep{merity2017pointer}. To verify that our findings transfer beyond curated Wikipedia text, we additionally run a subset of experiments on the first 1.5M-documents of FineWeb \citep{penedo2024fineweb} (\texttt{sample-10BT} configuration;
  $\approx$1\,B GPT-2 BPE tokens, with 1\% of documents held out for
  validation).

\paragraph{Conditioning.}

We use the conditioning set distribution described in Section 3 with $r_{\min} = 0$ and $b_{\min} = 1$, $b_{\max} = |\mathbf{x}_c|$, and sweep $r_{\max} \in {0.2, 0.4, 0.6, 0.8, 1.0}$ across experiments. Setting $r_{\min} = 0$ allows the conditioning set to be empty, in whichz case the model reduces to standard left-to-right language modeling. All non-conditioning tokens are included in the evaluation set, i.e., $\mathbf{x}_e = \mathbf{x} \setminus \mathbf{x}_c$.

\paragraph{Evaluation.}

We evaluate \ours{} across multiple modes, illustrated with a concrete example in \cref{eval_visualize}.
\begin{itemize}[leftmargin=*]
    \item \textit{Unconditional}: Empty conditioning set, perplexity over the entire sequence.
    \item \textit{Training distribution}: Conditioning sets sampled from the same distribution used at training time, perplexity over the evaluation set.
    \item \textit{Training distribution (no future)}: Same as above, except we do not condition a given token on anything that is in its future. This amounts to an empty conditioning set, but where we only report perplexity on tokens that are in the evaluation set of the \textit{Training distribution} mode. This mode is meant to assess the improved performance gained by conditioning on future tokens, which standard causal Transformers cannot do.
    \item \textit{Infilling}: Given a sequence $x=(x_1,\ldots,x_T)$, we partition it into three segments: a left boundary $l=(x_1,\ldots,x_l)$, a middle region $m=(x_{l+1},\ldots,x_{T-r})$, and a right boundary $r=(x_{T-r+1},\ldots,x_T)$. We define the conditioning set as the union of the two boundaries, $\mathbf{x}_c = l \cup r$, and the evaluation set as the middle region, $\mathbf{x}_e = m$. Following the procedure above, after sampling the total conditioning set size $|\mathbf{x}_c|$ from our conditioning distribution, we sample the fraction of conditioning tokens allocated to the left boundary, $|l|/(|l|+|r|) \sim \mathrm{Unif}(f_{\min}, f_{\max})$, and set $|r| = |\mathbf{x}_c| - |l|$. In our experiments, we fix $(f_{\min}, f_{\max})=(0.2, 0.8)$.
    This yields a conditioning set consisting of two contiguous blocks anchored at the beginning and end of the sequence. We report perplexity computed only over the middle region $\mathbf{x}_e$. This mode is meant to assess performance on conditional queries that could subserve a realistic downstream task.
    \item \textit{Infilling (no future)}: Same as above, except we do not condition a given token on anything that is in its future.
\end{itemize}

For all modes, we report the mean per-token perplexity within the evaluation set $\mathbf{x}_e$. In addition to likelihood evaluation, we also evaluate \textit{generation} for all of the above modes. Generation is done by sampling the evaluation set tokens one at a time conditioned on $\mathbf{x}_c$, using nucleus (top-$p$) sampling \citep{holtzman2019curious}. To score the quality of these generations, we use the established MAUVE score \citep{pillutla2021mauve}, which measures the gap between the generated and reference text distributions via a divergence frontier computed in a frozen embedding space, and has been shown to correlate well with human judgments of open-ended generation quality. See \cref{app:generation_details} for further details on how samples were generated and scored.

\subsection{Training Models From Scratch}
\label{sec:results}

\begin{figure*}[ht]
    \centering
    \includegraphics[width=\linewidth]{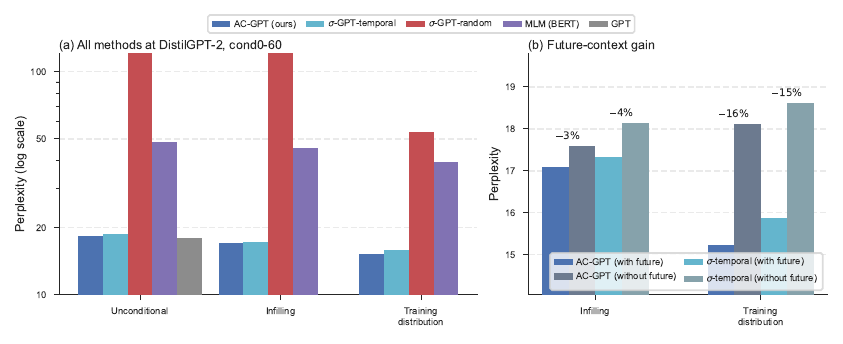}
    \caption{\textbf{Main results} ($r_{\max} = 0.6$, GPT-2 Small (82M)). \textbf{(a)} Perplexity ($\downarrow$) across evaluation modes. \textbf{(b)} Training-distribution perplexity ($\downarrow$) with vs.\ without future context, for \ours{} and \sigmagpt{}-temporal. The other methods reported in full in \cref{app:ablation}.}
    \label{fig:main_results}
\end{figure*}

We train a GPT-2-based architecture \citep{radford2019language} from
scratch following the datasets and conditioning protocol described in \cref{sec:experimental_setup}.We report results using GPT-2 Small (82M) at $r_{\max} = 0.6$ as the representative configuration; full ablations across model scales and conditioning ranges are in \cref{app:ablation}. We summarize the main results in \cref{fig:main_results}.

\paragraph{Comparison Methods.}

We compare \ours{} against the following baselines.

\textbf{\gpt}: A standard causal Transformer, which we can only evaluate on modes that involve left-to-right conditioning. The purpose of this baseline is to see if \ours{} degrades performance on left-to-right conditioning in virtue of having to model arbitrary conditionals.
    
\textbf{\sigmagpt{}}: The conditioning set is placed at the start of the sequence and the evaluation set is placed afterwards. For fair comparison to our model, we only backpropagate loss for prediction errors in the evaluation set, despite the fact that this is not strictly necessary for \sigmagpt{}. We note that if prediction losses for all tokens were backpropagated, it would speed up training, but would not change final model performance. We compare against two formulations of \sigmagpt{}.
\begin{enumerate}
    \item \textbf{\sigmagpt{}-random} faithfully implements the original \sigmagpt{} formulation \citep{pannatier2024sigma}: it randomly scrambles the orderings of tokens within the conditioning and evaluation sets, thereby coupling arbitrary conditionals with arbitrary token orderings.
    \item \textbf{\sigmagpt{}-temporal} is a controlled variant that we introduce to test our central hypothesis regarding the benefits of modeling arbitrary conditionals under a single left-to-right decomposition. \sigmagpt{}-temporal preserves \sigmagpt{}'s architecture (double positional encoding included) but constrains the orderings on which it is trained to be exclusively left-to-right within both the conditioning and evaluation sets. By design, \sigmagpt{}-temporal isolates the effect of decoupling within-set ordering from \sigmagpt{}'s broader architectural commitments. Under our hypothesis, \sigmagpt{}-temporal should escape the capacity penalty incurred by \sigmagpt{}-random and approach the performance of \ours{}.
\end{enumerate}

\textbf{\mlm{}}: The conditioning set is observed, while the evaluation set is masked and decoded. During training, we unmask the entire evaluation set in parallel to maintain efficiency, which maximizes the marginal likelihood of each token independently. At inference, we compute joint likelihoods by unmasking the evaluation set one token at a time from left to right using multiple forward passes.

\textbf{\diffusion{}}: The conditioning set is observed, while the evaluation set is masked and decoded. Given that diffusion models cannot evaluate likelihoods, we only score their sampling abilities.

For \ours{} and for all baselines, we build off of a base GPT-2 architecture \citep{radford2019language} and make only minimal modifications necessary to implement each method (\eg, for \sigmagpt{} we use double positional encodings, for \mlm{} we do not use causal attention). We use the same conditioning set distribution to train all models, for fair comparison.

\paragraph{Both \ours{} and \sigmagpt{}-temporal validate our central hypothesis.}

Among the methods evaluated, only \ours{} and \sigmagpt{}-temporal preserve a vanilla \gpt{}'s performance on \textit{Unconditional} generation (perplexity of 18.4 and 18.8 vs.\ 17.9) and demonstrate the best performance on conditional queries. \mlm{} (BERT) achieves a perplexity 2--3$\times$ worse than \ours{} across evaluation modes (48.4 vs.\ 18.4 on \textit{Unconditional}, 45.4 vs.\ 17.1 on \textit{Infilling}). \sigmagpt{}-random fails catastrophically across all modes. The convergence of \ours{} and \sigmagpt{}-temporal --- despite their architectural differences --- directly validates our central hypothesis: decoupling the modeling of arbitrary ordering fundamentally helps with modeling arbitrary conditionals. Both methods retain competitive perplexity on FineWeb (\cref{app:fineweb_likelihood}) and comparable sampling quality on these two datasets (\cref{app:generation_details}), confirming their viability extends beyond curated Wikipedia text.

\paragraph{Both methods leverage future conditioning.}

To isolate the contribution of future tokens, we compare each model's \textit{Infilling} and \textit{Training distribution} perplexities with and without access to future context: on \textit{Infilling}, \ours{} reduces perplexity from 17.6 to 17.1 (a 3\% reduction) and \sigmagpt{}-temporal from 18.1 to 17.3 (4\%); on \textit{Training distribution}, \ours{} reduces perplexity from 18.1 to 15.2 (a 16\% reduction) and \sigmagpt{}-temporal from 18.6 to 15.9 (15\%). Both methods derive comparable benefit from future conditioning, suggesting that the value of arbitrary conditional modeling on natural languages is paradigm-level (disentangling it from arbitrary token orderings) rather than tied to a specific architecture.

\paragraph{\ours{}'s unique advantages: scaling up through minimal architectural changes.}

With a simpler architecture than \sigmagpt{}-temporal, \ours{} achieves comparable perplexity when trained from scratch. Crucially, this same simplicity is the key to scale: \ours{} reuses the standard causal Transformer architecture exactly, enabling direct fine-tuning of pretrained large language models at billion-parameter scale (\cref{sec:fine-tuning}). \sigmagpt{}-temporal, by contrast, requires substantial architectural surgery to existing checkpoints, foreclosing this scaling pathway.

\subsection{Fine-Tuning Large Language Models}
\label{sec:fine-tuning}

\begin{figure}[ht]
    \centering
    \includegraphics[width=\linewidth]{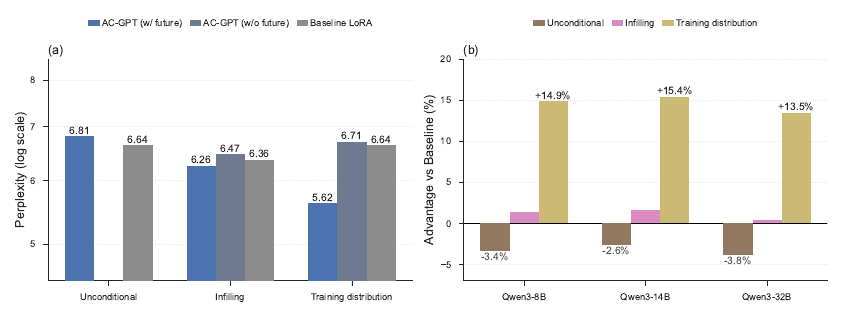}
    \caption{\textbf{(a)} Qwen3-14B + LoRA $r=8$, $r_{\max} = 0.6$. Perplexity ($\downarrow$) for AC-GPT LoRA with and without future context, compared to a standard causal LoRA baseline. \textbf{(b)} Relative perplexity advantage (\%) of AC-GPT LoRA over the causal LoRA baseline at $r_{\max}=0.6$ across three Qwen3 scales (LoRA $r=8$); }
\label{fig:finetune_results}
\end{figure}

To verify that \ours{} scales to pretrained large language models, we conduct fine-tuning experiments along several axes: a backbone scaling sweep, cross-family transfer, an adapter-capacity comparison (LoRA vs.\ full fine-tuning), training-corpus generalization, and a conditioning-range sweep. Throughout this section, we compare the results on the five evaluation modes from \cref{sec:experimental_setup}; \cref{fig:finetune_results} reports the headline results and \cref{appendix:finetune-extras} reports per-cell numbers for every configuration.

\paragraph{\ours{} achieves substantial gains through finetuning.}

We highlight Qwen3-14B with $r_{\max}=0.6$ as a representative configuration. \ours{} LoRA achieves a \textit{Training-distribution} perplexity of 5.62 with future context, outperforming the causal baseline (6.64) by 15.4\%. Without future context, \ours{} matches the baseline on Training distribution (6.71 vs.\ 6.64), confirming the gain comes entirely from leveraging future tokens rather than from any difference in left-to-right modeling. On \textit{Infilling}, \ours{} with future context (6.26) slightly outperforms the baseline (6.36), offering fewer benefits; on \textit{Unconditional}, the model pays a small perplexity cost (6.81 vs.\ 6.64), a favorable trade for the 15.4\% gain on conditional queries that exploit future context --- a capability the causal baseline lacks by construction. Smaller gains on infilling might be due to differences between the infilling setting and the training distribution over conditional queries, and larger gains could potentially be obtained by adjusting the hyperparameters of the conditional set distribution described in \cref{sec:methods} to target infilling as a downstream task. 

\paragraph{The gains hold across model scales.}

This pattern persists as we vary the backbone size: as shown in \cref{fig:finetune_results}b, the \textit{Training-distribution} gain stays in the 13.5--15.4\% range across all three Qwen3 backbones (8B, 14B, 32B), and the \textit{Unconditional} cost is similarly uniform ($-2.5\%$ to $-3.9\%$). Notably, the gain magnitude does not collapse at the 32B scale, indicating that pretrained models with stronger left-to-right priors do not erase the benefit of future-context conditioning. The consistency across an order-of-magnitude range in parameter count suggests the gain is a structural property of the conditioning set rather than an artifact of any specific scale.

\paragraph{The gain transfers across family, capacity, and corpus; magnitude scales with conditioning range.}

These results are not specific to Qwen3. Repeating the same fine-tuning
protocol on LLaMA-3.1-8B with a higher-capacity adapter (LoRA $r=64$) yields a \textit{Training-distribution} advantage of $+15.3\%$, indistinguishable from the Qwen3 results despite the $8\times$ difference in adapter capacity. Replacing LoRA with full fine-tuning on the same backbone produces nearly identical perplexity (5.13 full FT vs.\ 5.11 LoRA $r=64$, a 0.4\% gap despite a roughly $50\times$ difference in trainable parameters), ruling out adapter expressiveness as the source of the gain. Switching the training corpus from Wikitext-103 to FineWeb leaves the advantage essentially unchanged ($+16.9\%$ vs.\ $+15.3\%$ on the same LLaMA-3.1-8B backbone), indicating the gain is not specific to a small encyclopedic corpus. On Qwen3-1.7B with the same LoRA configuration, sweeping the conditioning range $r_{\max} \in \{0.2, 0.4, 0.6, 0.8, 1.0\}$ yields a \textit{Training-distribution} advantage that scales monotonically from $+1.2\%$ to $+11.6\%$, with our primary $r_{\max}=0.6$ yielding $+7.4\%$ as a balance between conditioning breadth and capability concentration. Per-cell numbers, AR-mixing variants, and the full $r_{\max}$ sweep are reported in \cref{appendix:finetune-extras}.

\paragraph{Modular deployment via mode-specific adapters.}

The near-equivalence between LoRA and full fine-tuning shown above indicates that the \ours{} objective can be fine-tuned at LoRA rank with negligible performance loss. This invites a deployment-friendly extension we leave to future work: training mode-specific adapters (\eg, one optimized for \textit{Infilling}, one for \textit{Training distribution}) on a shared frozen backbone, switched at inference time according to the query type. Such a setup amortizes a single pretraining cost across multiple specialized conditional behaviors.

\section{Conclusion}
\label{sec:conclusion}

In this work, we introduced \ours{}, a modification to standard causal Transformers that enables the evaluation and sampling of arbitrary conditionals. By augmenting the context with position-aware conditioning tokens, our method preserves the left-to-right decomposition and next-token prediction objective essential for modeling natural language. 
Our results demonstrate that \ours{} outperforms existing baselines on arbitrary-conditional approaches without sacrificing performance on standard left-to-right ones, and uniquely enables direct fine-tuning of pretrained LLMs for arbitrary conditional tasks at billion-parameter scale.

We focused our empirical analysis on queries in which each token is either in the evaluation set or conditioning set, but note that this omits queries that marginalize over an \textit{unknown set} of tokens. This can easily be achieved in \ours{} by introducing an additional \texttt{MASK} token to replace the tokens in the unknown set. In this work, we prioritized conditioning because it addresses the fundamental inability of causal models to incorporate future context, whereas marginalization is technically feasible in standard architectures through the use of an additional \texttt{MASK} token as well. We leave exploration of marginalization queries to future work.

We acknowledge two limitations in the scope of our work. First, \ours{} trades training efficiency for architectural compatibility with pretrained models: compared to any-order models like \sigmagpt{} which can compute loss on all tokens simultaneously, \ours{} only backpropagates loss through the designated evaluation set. This trade-off is favorable in the practical setting where a pretrained LLM is already available, but disadvantageous for from-scratch training under tight compute budgets. Second, our experiments were restricted to two datasets, albeit fairly large ones. Future work should extend this evaluation to even larger corpora and explore specific downstream applications such as infilling and constrained text editing.

\begin{ack}
We thank Dhanya Sridhar for helping with ideation at the beginning of the project.
EE acknowledges support from Vanier Canada Graduate Scholarship \#492702.
GL acknowledges support from NSERC Discovery Grant RGPIN-2018-04821, the Canada Research Chair in Neural Computations and Interfacing, and a Canada-CIFAR AI Chair.
\end{ack}

\bibliography{references}
\bibliographystyle{apalike}

\begin{appendices}
\onecolumn
\crefalias{section}{appendix}   
\crefalias{subsection}{appendix} 
\counterwithin{figure}{section} 
\counterwithin{table}{section}  
\clearpage

\section{Details for Sampling Experiments}

\paragraph{Sample generation.}
For each of the five evaluation modes from \cref{sec:experimental_setup} we generate $n_p$ samples per model per random seed ($n_p = 1000$ for FineWeb sample-10BT and $n_p = 241$ for Wikitext-103; three seeds total). The conditioning range $r_{\max}$ used at sampling time is reported in each table's caption. All methods share a single sampling specification: nucleus sampling \citep{holtzman2019curious} with $p = 0.95$ and temperature $T = 0.8$. For \textit{Unconditional} generation each sample is conditioned on a fixed $20$-token prompt drawn from the validation split of the corpus on which the model was trained, after which the model autoregressively produces $1004$ further tokens, for a total sequence length of $1024$. For the four conditional modes, the conditioning set $\mathbf{x}_c$ is taken from a pre-built evaluation configuration shared across methods, so for any given table cell all methods condition on identical $\mathbf{x}_c$; each model then samples $\mathbf{x}_e$ one token at a time conditioned on $\mathbf{x}_c$ following its native decoding procedure. The \textit{Infilling} and \textit{Training distribution} modes reuse the same conditioning sets as in the likelihood evaluation, whereas the two no-future modes are simulated with the \textit{Unconditional} procedure, generating left to right while fixing each conditioning-set position to its ground-truth token so that no token is conditioned on future context.

\paragraph{MAUVE scoring.}
We score generation quality with MAUVE \citep{pillutla2021mauve} against an in-distribution reference distribution $Q$ drawn from the same corpus on which the model under evaluation was trained. The reference is constructed once per corpus and cached. For Wikitext-103, we walk the test split paragraph by paragraph (skipping section headers and blank entries) and retain the first $n_q = 200$ paragraphs whose GPT-2 large tokenization is at least $1024$ tokens, truncating each to the first $1024$ tokens. For FineWeb sample-10BT, we read the tokenized validation \texttt{.bin}, split it on the GPT-2 end-of-text token, and retain the first $n_q = 1000$ documents that are at least $1024$ tokens long and begin at a token offset of at least $1{,}024{,}000$, ensuring the reference is disjoint from the validation region from which prompts and conditioning contexts are drawn; each document is truncated to its first $1024$ tokens. Generated samples (the distribution $P$) are likewise truncated to the first $1024$ tokens. Both $P$ and $Q$ are featurized into $1280$-dimensional vectors by reading the GPT-2 large \citep{radford2019language} hidden state at the last non-padding position of each sequence. MAUVE then computes the divergence frontier between $P$ and $Q$ via $k$-means quantization (a maximum of $500$ iterations, $5$ random restarts, with the best-loss restart selected). We report the mean and standard deviation of the resulting score across the three model-training seeds.

\paragraph{Scope of MAUVE on conditional modes.}
For the four conditional modes, MAUVE is computed over the full reconstructed sequence ($\mathbf{x}_c \cup \mathbf{x}_e$) rather than the generated portion $\mathbf{x}_e$ alone. Because the conditioning tokens are identical across methods within any given table cell, this contributes a shared positive bias toward $Q$ that is the same for all compared methods, so the cross-method comparisons reported in our MAUVE tables are unaffected. The absolute MAUVE values nonetheless overstate the distributional similarity that would be observed if scoring were restricted to the generated tokens in isolation.

\paragraph{Takeaways.}
Across the WikiText-103 and FineWeb experiments, the sampling results mirror the likelihood results in several respects. AC-GPT and $\sigma$-GPT-temporal again track each other closely and are the strongest of the methods that condition on future context, most clearly in the Training-distribution mode, while on Unconditional generation both remain on par with the vanilla GPT baseline. As with likelihood, the gap between Training-distribution generation with and without future tokens reflects the ability of AC-GPT and $\sigma$-GPT-temporal to utilize future information.

On WikiText-103 a 20-token prompt is already enough for both models to reach high Unconditional MAUVE. On FineWeb, a noisier corpus on which unconditional generation from only a 20-token prompt degenerates under our decoding hyperparameters for every method, including the GPT baseline, AC-GPT and $\sigma$-GPT-temporal instead recover high-quality generation by conditioning on the scattered known tokens of the Training-distribution mode, where their MAUVE climbs from near the floor back to roughly its WikiText level. This recovery is exactly what $\sigma$-GPT-random cannot achieve.

We use MAUVE to corroborate our claims but treat it as supporting evidence rather than a fully reliable measure of generation quality. In the Infilling modes in particular it is dominated by the long contiguous ground-truth boundaries and is largely insensitive to the generated span, so every method scores highly; the qualitative samples (\cref{app:gen-samples}) nonetheless show $\sigma$-GPT-random collapsing into incoherent word salad there, while AC-GPT and $\sigma$-GPT-temporal stay fluent.

\label{app:generation_details}

\subsection{Sampling ablation results}\label{app:sample-ablation}

This part reports our \emph{sampling ablation}: for each method we generate samples under every evaluation mode and conditioning range, and score their quality with MAUVE ($\uparrow$; \citealp{pillutla2021mauve}) against an in-distribution reference. \Cref{tab:mauve_small,tab:mauve_base,tab:mauve_medium} cover WikiText-103 at the Small (82M), Base (124M), and Medium (204M) backbone sizes; \cref{tab:fineweb-mauve-pl20} covers FineWeb sample-10BT. Bold marks the best method within each column.We use a prompt length of 20 for the unconditional method as mentioned above. 
\input{sampling_ablation_table_1024}
\begin{table}[t]
\caption{\textbf{Generation quality (MAUVE$\uparrow$) on FineWeb across the five evaluation modes (\cref{fig:eval_modes})} We use a prompt length of 20 for the unconditional method. GPT-2 base (124M) models trained from scratch on FineWeb sample-10BT, evaluated at $r_{\max} = 0.4$; mean $\pm$ std over three seeds. ``--'' marks modes that require future-context conditioning, which the causal baseline cannot perform. Bold = column-wise best.}
\label{tab:fineweb-mauve-pl20}
\begin{center}
\resizebox{\columnwidth}{!}{%
\begin{tabular}{lccccc}
\toprule
Model              & Unconditional              & Infilling                  & Training dist.             & Infilling (no future)      & Training dist. (no future) \\
\midrule
AC-GPT             & $0.085 \pm 0.017$          & $0.934 \pm 0.025$          & $\mathbf{0.893 \pm 0.009}$ & $0.934 \pm 0.030$          & $0.337 \pm 0.038$          \\
$\sigma$-temporal  & $0.067 \pm 0.020$          & $\mathbf{0.951 \pm 0.005}$ & $0.859 \pm 0.031$          & $\mathbf{0.955 \pm 0.013}$ & $\mathbf{0.373 \pm 0.112}$ \\
$\sigma$-random    & $0.006 \pm 0.001$          & $0.815 \pm 0.022$          & $0.008 \pm 0.001$          & $0.778 \pm 0.043$          & $0.007 \pm 0.000$          \\
GPT                & $\mathbf{0.101 \pm 0.027}$ & --                         & --                         & $0.923 \pm 0.015$          & $0.355 \pm 0.021$          \\
\bottomrule
\end{tabular}%
}
\end{center}
\end{table}

\clearpage
\input{generation_samples}
\clearpage
\section{Ablation Results}
\label{app:ablation}
This appendix reports the full likelihood ablation behind the main results: perplexity ($\downarrow$) for every method across the five evaluation modes and five conditioning ranges ($r_{\max}$ from $0.2$ to $1.0$). \Cref{tab:ablation_small,tab:ablation_base,tab:ablation_medium} cover WikiText-103 at the Small (82M), Base (124M), and Medium (204M) backbone sizes; the corresponding FineWeb sample-10BT evaluation is in \cref{tab:fineweb_likelihood}.

\input{ablation_table.tex}

\clearpage
\subsection{FineWeb Likelihood Evaluation}
\label{app:fineweb_likelihood}

\begin{table}[ht]
\caption{\textbf{FineWeb likelihood evaluation.} GPT-2 base (124M) models trained from scratch on FineWeb sample-10BT, evaluated at $r_{\max} = 0.4$; mean $\pm$ std over three seeds. ``--'' marks modes that require future-context conditioning, which the causal baseline cannot perform. Bold = column-wise best.}
\label{tab:fineweb_likelihood}
\begin{center}
\resizebox{\columnwidth}{!}{%
\begin{tabular}{lccccc}
\toprule
Model              & Unconditional             & Infilling                 & Training dist.            & Infilling (no future)     & Training dist. (no future) \\
\midrule
AC-GPT             & $28.79 \pm 0.08$          & $\mathbf{26.74 \pm 0.04}$ & $\mathbf{24.06 \pm 0.03}$ & $27.70 \pm 0.02$          & $28.60 \pm 0.07$           \\
$\sigma$-temporal  & $29.00 \pm 0.14$          & $26.94 \pm 0.05$          & $24.82 \pm 0.05$          & $28.08 \pm 0.13$          & $28.94 \pm 0.13$           \\
$\sigma$-random    & $593.46 \pm 224.63$       & $805.77 \pm 370.95$       & $51.30 \pm 0.59$          & $650.26 \pm 304.83$       & $602.08 \pm 236.37$        \\
GPT                & $\mathbf{27.58 \pm 0.04}$ & --                        & --                        & $\mathbf{26.85 \pm 0.04}$ & $\mathbf{27.48 \pm 0.05}$  \\
\bottomrule
\end{tabular}%
}
\end{center}
\end{table}

\clearpage
\section{Additional fine-tuning results}
\label{appendix:finetune-extras}
\begin{figure}[ht]
  \centering
  \includegraphics[width=\linewidth]{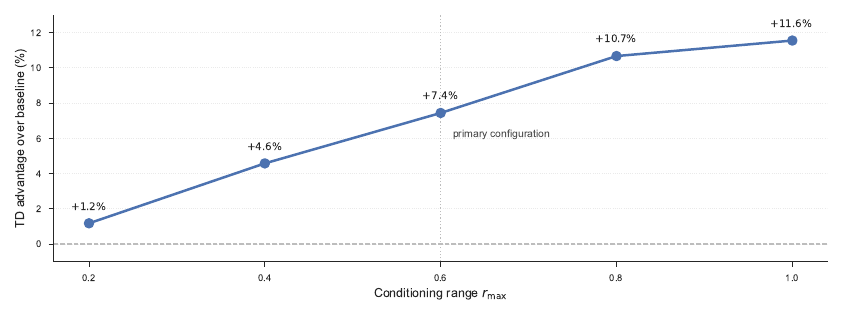}
  \caption{Qwen3-1.7B $r_{\max}$ sweep (LoRA $r=8$, Wikitext-103). Relative \textit{Training-distribution} perplexity advantage (\%) of \ours{} over the matched-budget causal LoRA baseline as a function of conditioning range. Advantage scales monotonically with $r_{\max}$.}
  \label{fig:rmax-sweep}
\end{figure}

This appendix gives the per-configuration results behind the fine-tuning study of \cref{sec:fine-tuning}. We fine-tune Qwen3-8B, Qwen3-14B, Qwen3-32B, and LLaMA-3.1-8B with LoRA (plus a full fine-tuning run for LLaMA-3.1-8B) on WikiText-103 and FineWeb, varying the conditioning range and the autoregressive-mixing probability. \Cref{tab:finetune-ppls} lists absolute perplexities, \cref{tab:finetune-adv} the relative advantage of AC-GPT over the matched-budget causal baseline, and \cref{fig:rmax-sweep} the effect of the conditioning range on a Qwen3-1.7B sweep.

\begin{table}[ht]
\caption{Per-configuration perplexities for the fine-tuning sweep (\cref{sec:fine-tuning}). All cells use $r_{\max}=0$ (causal baseline) or $r_{\max}=0.6$ (AC-GPT). \emph{Mix} denotes the pure-autoregressive mixing probability during training (0 means no mixing). Mode definitions follow \cref{sec:results}; the \textit{Infilling} and \textit{Training dist.} columns evaluate AC-GPT with future context, while \textit{Infilling w/o future} and \textit{Training dist.\ w/o future} evaluate without. Baseline cells report \texttt{N/A} for the with-future columns because a causal model cannot exploit future context.}
\label{tab:finetune-ppls}
\begin{center}
\resizebox{\columnwidth}{!}{%
\begin{tabular}{ll lll c c ccccc}
\toprule
Method & Backbone & Adapter & Corpus & $r_{\max}$ & Mix & Unconditional & Infilling & Training dist.\ & Infilling w/o future & Training dist.\ w/o future \\
\midrule
Baseline & Qwen3-8B & LoRA $r=8$ & Wikitext-103 & 0.0 & 0 & 7.25 & N/A & N/A & 6.95 & 7.25 \\
\midrule
AC-GPT & Qwen3-8B & LoRA $r=8$ & Wikitext-103 & 0.6 & 0 & 7.50 & 6.85 & 6.17 & 7.12 & 7.38 \\
AC-GPT & Qwen3-8B & LoRA $r=8$ & Wikitext-103 & 0.6 & 0.3 & 7.34 & 6.85 & 6.62 & 7.03 & 7.28 \\
\midrule
Baseline & Qwen3-14B & LoRA $r=8$ & Wikitext-103 & 0.0 & 0 & 6.64 & N/A & N/A & 6.36 & 6.64 \\
\midrule
AC-GPT & Qwen3-14B & LoRA $r=8$ & Wikitext-103 & 0.6 & 0 & 6.81 & 6.26 & 5.62 & 6.47 & 6.71 \\
AC-GPT & Qwen3-14B & LoRA $r=8$ & Wikitext-103 & 0.6 & 0.3 & 6.73 & 6.27 & 6.04 & 6.44 & 6.66 \\
\midrule
Baseline & Qwen3-32B & LoRA $r=8$ & Wikitext-103 & 0.0 & 0 & 6.35 & N/A & N/A & 6.09 & 6.35 \\
\midrule
AC-GPT & Qwen3-32B & LoRA $r=8$ & Wikitext-103 & 0.6 & 0 & 6.59 & 6.07 & 5.49 & 6.25 & 6.49 \\
AC-GPT & Qwen3-32B & LoRA $r=8$ & Wikitext-103 & 0.6 & 0.3 & 6.42 & 5.99 & 5.75 & 6.15 & 6.36 \\
\midrule
Baseline & Llama-3.1-8B & LoRA $r=64$ & Wikitext-103 & 0.0 & 0 & 6.04 & N/A & N/A & 5.78 & 6.04 \\
\midrule
AC-GPT & Llama-3.1-8B & LoRA $r=64$ & Wikitext-103 & 0.6 & 0 & 6.19 & 5.67 & 5.11 & 5.86 & 6.09 \\
AC-GPT & Llama-3.1-8B & LoRA $r=64$ & Wikitext-103 & 0.6 & 0.3 & 6.08 & 5.67 & 5.48 & 5.81 & 6.03 \\
AC-GPT & Llama-3.1-8B & Full FT & Wikitext-103 & 0.6 & 0 & 6.15 & 5.66 & 5.13 & 5.84 & 6.05 \\
\midrule
Baseline & Llama-3.1-8B & LoRA $r=64$ & FineWeb & 0.0 & 0 & 10.23 & N/A & N/A & 9.78 & 10.23 \\
\midrule
AC-GPT & Llama-3.1-8B & LoRA $r=64$ & FineWeb & 0.6 & 0 & 10.51 & 9.77 & 8.50 & 10.01 & 10.37 \\
AC-GPT & Llama-3.1-8B & LoRA $r=64$ & FineWeb & 0.6 & 0.1 & 10.38 & 9.58 & 8.64 & 9.91 & 10.26 \\
\bottomrule
\end{tabular}%
}
\end{center}
\end{table}

\begin{table}[ht]
\caption{Relative perplexity advantage (\%) of AC-GPT LoRA over the matched-budget causal LoRA baseline at $r_{\max}=0.6$. \emph{UC adv}: AC-GPT Unconditional vs.\ Baseline Unconditional. \emph{Inf adv}: AC-GPT Infilling (with future) vs.\ Baseline Infilling (w/o future). \emph{TD adv}: AC-GPT Training dist.\ (with future) vs.\ Baseline Training dist.\ (w/o future). Positive = AC-GPT better.}
\label{tab:finetune-adv}
\begin{center}
\resizebox{\columnwidth}{!}{%
\begin{tabular}{lll c rrr}
\toprule
Backbone & Adapter & Corpus & Mix & UC adv & Inf adv & TD adv \\
\midrule
Qwen3-8B & LoRA $r=8$ & Wikitext-103 & 0 & $-3.37\%$ & $+1.41\%$ & $+14.86\%$ \\
Qwen3-8B & LoRA $r=8$ & Wikitext-103 & 0.3 & $-1.27\%$ & $+1.45\%$ & $+8.65\%$ \\
Qwen3-14B & LoRA $r=8$ & Wikitext-103 & 0 & $-2.61\%$ & $+1.58\%$ & $+15.40\%$ \\
Qwen3-14B & LoRA $r=8$ & Wikitext-103 & 0.3 & $-1.35\%$ & $+1.49\%$ & $+9.02\%$ \\
Qwen3-32B & LoRA $r=8$ & Wikitext-103 & 0 & $-3.85\%$ & $+0.41\%$ & $+13.48\%$ \\
Qwen3-32B & LoRA $r=8$ & Wikitext-103 & 0.3 & $-1.14\%$ & $+1.65\%$ & $+9.45\%$ \\
Llama-3.1-8B & LoRA $r=64$ & Wikitext-103 & 0 & $-2.54\%$ & $+1.87\%$ & $+15.32\%$ \\
Llama-3.1-8B & LoRA $r=64$ & Wikitext-103 & 0.3 & $-0.66\%$ & $+1.96\%$ & $+9.18\%$ \\
Llama-3.1-8B\textsuperscript{*} & Full FT & Wikitext-103 & 0 & $-1.85\%$ & $+2.06\%$ & $+14.98\%$ \\
Llama-3.1-8B & LoRA $r=64$ & FineWeb & 0 & $-2.70\%$ & $+0.01\%$ & $+16.94\%$ \\
Llama-3.1-8B & LoRA $r=64$ & FineWeb & 0.1 & $-1.45\%$ & $+2.06\%$ & $+15.55\%$ \\
\bottomrule
\end{tabular}%
}
\\[3pt]
\footnotesize{\textsuperscript{*}\,No matched full-fine-tuning baseline was trained; the advantages reported for the LLaMA-3.1-8B Full FT row use the LoRA $r=64$ baseline on the same backbone and corpus as a reference.}
\end{center}
\end{table}

\clearpage
\section{Figures}

\begin{figure}[ht]
    \centering
    \includegraphics[width=\linewidth]{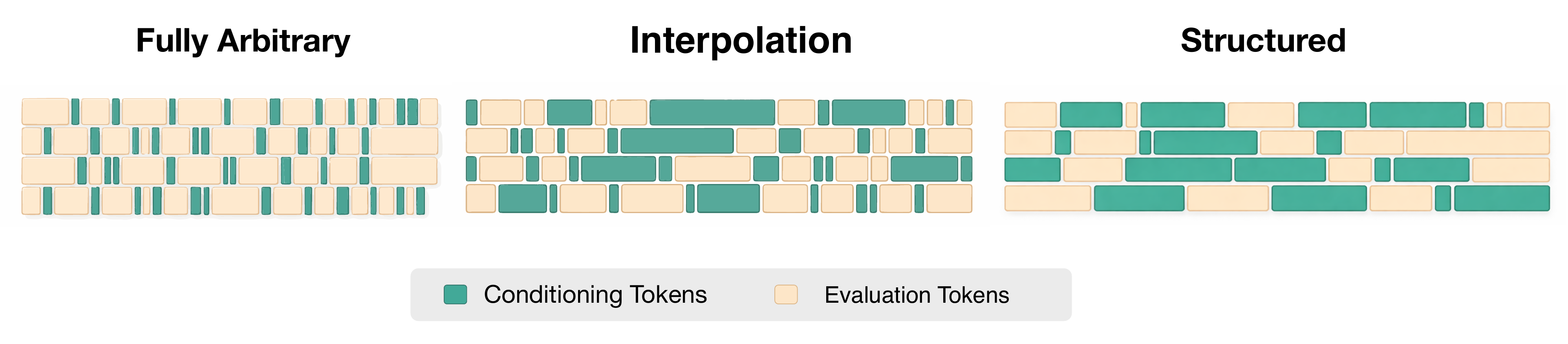}
    \caption{
    \textbf{Conditioning set construction via block-based sampling}.
    We illustrate the three regimes of selecting a conditioning set $\mathbf{x}_c$ from  a sequence of text $\mathbf{x}$: fully arbitrary, interpolated, and structured conditioning.
    Conditioning tokens (green) are selected as contiguous blocks, while remaining
    tokens (beige) are used for evaluation. Varying the number and sizes of blocks
    yields a smooth interpolation between full-arbitrary and structured conditioning. \textbf{Fully arbitrary} conditioning corresponds to many small, scattered conditioning blocks, yielding highly fragmented context.
\textbf{Structured conditioning} corresponds to a small number of long contiguous blocks, yielding coherent spans of context.
\textbf{Interpolation} smoothly bridges these extremes by varying both the number and sizes of contiguous conditioning blocks.
    }
    \label{fig:arbitrary-conditioning}
\end{figure}

\begin{figure}[ht]
\centering
\includegraphics[width=\textwidth]{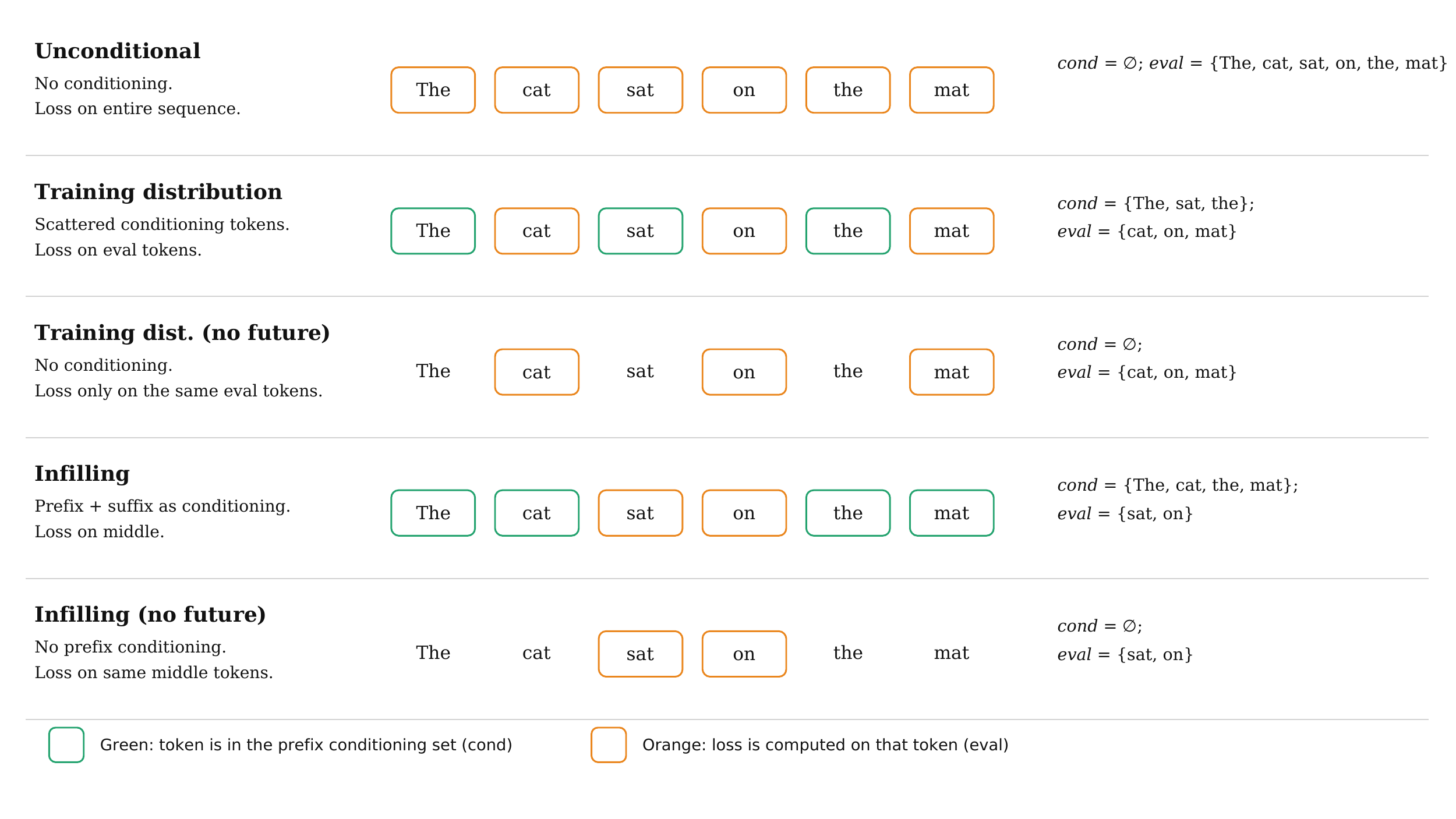}
\label{eval_visualize}
\caption{\textbf{The five evaluation modes used throughout the paper.} illustrated with ``The cat sat on the mat.'' as a running example. See\cref{sec:experimental_setup} for formal definitions.}
\label{fig:eval_modes}
\end{figure}
\end{appendices}

\newpage

\end{document}

%% file: sampling_ablation_table_1024.tex
\begin{table*}[ht]
\caption{MAUVE Ablation (Small): MAUVE ($\uparrow$) on WikiText-103 across evaluation modes and conditioning scales using GPT-2 Small backbone.}
\label{tab:mauve_small}
\begin{center}
\resizebox{\columnwidth}{!}{%
\begin{tabular}{ll ccccc}
\toprule
\textbf{Scale} & \textbf{Model} & \textbf{Unconditional} & \textbf{Infilling} & \textbf{Training dist.} & \textbf{Infilling w/o future} & \textbf{Training dist. w/o future} \\
\midrule
\multirow{6}{*}{cond0-20}
  & AC-GPT (ours) & 0.856 {\scriptsize $\pm$ 0.019} & 0.936 {\scriptsize $\pm$ 0.014} & \textbf{0.944 {\scriptsize $\pm$ 0.009}} & 0.934 {\scriptsize $\pm$ 0.041} & 0.817 {\scriptsize $\pm$ 0.049} \\
  & $\sigma$-GPT-temporal & 0.867 {\scriptsize $\pm$ 0.090} & \textbf{0.964 {\scriptsize $\pm$ 0.012}} & 0.923 {\scriptsize $\pm$ 0.037} & \textbf{0.962 {\scriptsize $\pm$ 0.009}} & 0.829 {\scriptsize $\pm$ 0.107} \\
  & $\sigma$-GPT-random & 0.007 {\scriptsize $\pm$ 0.001} & 0.666 {\scriptsize $\pm$ 0.023} & 0.044 {\scriptsize $\pm$ 0.014} & 0.759 {\scriptsize $\pm$ 0.013} & 0.018 {\scriptsize $\pm$ 0.003} \\
  & MLM (BERT) & 0.159 {\scriptsize $\pm$ 0.078} & 0.877 {\scriptsize $\pm$ 0.042} & 0.373 {\scriptsize $\pm$ 0.195} & 0.809 {\scriptsize $\pm$ 0.007} & 0.207 {\scriptsize $\pm$ 0.098} \\
  & Diffusion & 0.165 {\scriptsize $\pm$ 0.036} & 0.472 {\scriptsize $\pm$ 0.032} & 0.344 {\scriptsize $\pm$ 0.031} & 0.889 {\scriptsize $\pm$ 0.038} & 0.445 {\scriptsize $\pm$ 0.033} \\
  & GPT & \textbf{0.923 {\scriptsize $\pm$ 0.046}} & -- & -- & 0.944 {\scriptsize $\pm$ 0.017} & \textbf{0.873 {\scriptsize $\pm$ 0.087}} \\
\cmidrule{1-7}
\multirow{6}{*}{cond0-40}
  & AC-GPT (ours) & 0.869 {\scriptsize $\pm$ 0.092} & 0.938 {\scriptsize $\pm$ 0.024} & \textbf{0.943 {\scriptsize $\pm$ 0.015}} & 0.918 {\scriptsize $\pm$ 0.031} & 0.816 {\scriptsize $\pm$ 0.145} \\
  & $\sigma$-GPT-temporal & 0.878 {\scriptsize $\pm$ 0.033} & \textbf{0.950 {\scriptsize $\pm$ 0.034}} & 0.931 {\scriptsize $\pm$ 0.032} & \textbf{0.963 {\scriptsize $\pm$ 0.016}} & 0.771 {\scriptsize $\pm$ 0.026} \\
  & $\sigma$-GPT-random & 0.014 {\scriptsize $\pm$ 0.003} & 0.866 {\scriptsize $\pm$ 0.108} & 0.151 {\scriptsize $\pm$ 0.113} & 0.864 {\scriptsize $\pm$ 0.058} & 0.023 {\scriptsize $\pm$ 0.010} \\
  & MLM (BERT) & 0.296 {\scriptsize $\pm$ 0.172} & 0.921 {\scriptsize $\pm$ 0.039} & 0.698 {\scriptsize $\pm$ 0.070} & 0.906 {\scriptsize $\pm$ 0.039} & 0.327 {\scriptsize $\pm$ 0.117} \\
  & Diffusion & 0.100 {\scriptsize $\pm$ 0.013} & 0.712 {\scriptsize $\pm$ 0.022} & 0.584 {\scriptsize $\pm$ 0.200} & 0.921 {\scriptsize $\pm$ 0.059} & 0.426 {\scriptsize $\pm$ 0.052} \\
  & GPT & \textbf{0.923 {\scriptsize $\pm$ 0.046}} & -- & -- & 0.933 {\scriptsize $\pm$ 0.008} & \textbf{0.842 {\scriptsize $\pm$ 0.075}} \\
\cmidrule{1-7}
\multirow{6}{*}{cond0-60}
  & AC-GPT (ours) & 0.883 {\scriptsize $\pm$ 0.068} & 0.926 {\scriptsize $\pm$ 0.016} & 0.914 {\scriptsize $\pm$ 0.017} & 0.908 {\scriptsize $\pm$ 0.064} & 0.806 {\scriptsize $\pm$ 0.092} \\
  & $\sigma$-GPT-temporal & 0.806 {\scriptsize $\pm$ 0.051} & 0.949 {\scriptsize $\pm$ 0.023} & \textbf{0.942 {\scriptsize $\pm$ 0.009}} & \textbf{0.956 {\scriptsize $\pm$ 0.025}} & 0.669 {\scriptsize $\pm$ 0.079} \\
  & $\sigma$-GPT-random & 0.012 {\scriptsize $\pm$ 0.002} & \textbf{0.952 {\scriptsize $\pm$ 0.020}} & 0.144 {\scriptsize $\pm$ 0.093} & 0.898 {\scriptsize $\pm$ 0.046} & 0.036 {\scriptsize $\pm$ 0.004} \\
  & MLM (BERT) & 0.431 {\scriptsize $\pm$ 0.181} & 0.885 {\scriptsize $\pm$ 0.054} & 0.756 {\scriptsize $\pm$ 0.065} & 0.927 {\scriptsize $\pm$ 0.026} & 0.315 {\scriptsize $\pm$ 0.156} \\
  & Diffusion & 0.120 {\scriptsize $\pm$ 0.036} & 0.836 {\scriptsize $\pm$ 0.010} & 0.671 {\scriptsize $\pm$ 0.104} & 0.942 {\scriptsize $\pm$ 0.021} & 0.508 {\scriptsize $\pm$ 0.137} \\
  & GPT & \textbf{0.923 {\scriptsize $\pm$ 0.046}} & -- & -- & 0.947 {\scriptsize $\pm$ 0.009} & \textbf{0.874 {\scriptsize $\pm$ 0.117}} \\
\cmidrule{1-7}
\multirow{6}{*}{cond0-80}
  & AC-GPT (ours) & 0.805 {\scriptsize $\pm$ 0.061} & 0.914 {\scriptsize $\pm$ 0.017} & \textbf{0.947 {\scriptsize $\pm$ 0.030}} & 0.946 {\scriptsize $\pm$ 0.013} & 0.767 {\scriptsize $\pm$ 0.128} \\
  & $\sigma$-GPT-temporal & 0.836 {\scriptsize $\pm$ 0.094} & \textbf{0.964 {\scriptsize $\pm$ 0.008}} & 0.927 {\scriptsize $\pm$ 0.023} & 0.923 {\scriptsize $\pm$ 0.017} & 0.786 {\scriptsize $\pm$ 0.087} \\
  & $\sigma$-GPT-random & 0.011 {\scriptsize $\pm$ 0.001} & 0.962 {\scriptsize $\pm$ 0.006} & 0.284 {\scriptsize $\pm$ 0.228} & \textbf{0.949 {\scriptsize $\pm$ 0.025}} & 0.059 {\scriptsize $\pm$ 0.033} \\
  & MLM (BERT) & 0.227 {\scriptsize $\pm$ 0.041} & 0.947 {\scriptsize $\pm$ 0.016} & 0.850 {\scriptsize $\pm$ 0.012} & 0.897 {\scriptsize $\pm$ 0.024} & 0.335 {\scriptsize $\pm$ 0.024} \\
  & Diffusion & 0.084 {\scriptsize $\pm$ 0.031} & 0.856 {\scriptsize $\pm$ 0.033} & 0.760 {\scriptsize $\pm$ 0.043} & 0.925 {\scriptsize $\pm$ 0.018} & 0.476 {\scriptsize $\pm$ 0.082} \\
  & GPT & \textbf{0.923 {\scriptsize $\pm$ 0.046}} & -- & -- & 0.914 {\scriptsize $\pm$ 0.016} & \textbf{0.791 {\scriptsize $\pm$ 0.031}} \\
\cmidrule{1-7}
\multirow{6}{*}{cond0-100}
  & AC-GPT (ours) & 0.844 {\scriptsize $\pm$ 0.030} & \textbf{0.942 {\scriptsize $\pm$ 0.025}} & \textbf{0.963 {\scriptsize $\pm$ 0.022}} & 0.951 {\scriptsize $\pm$ 0.033} & 0.808 {\scriptsize $\pm$ 0.119} \\
  & $\sigma$-GPT-temporal & 0.664 {\scriptsize $\pm$ 0.067} & 0.919 {\scriptsize $\pm$ 0.032} & 0.914 {\scriptsize $\pm$ 0.038} & 0.950 {\scriptsize $\pm$ 0.038} & 0.683 {\scriptsize $\pm$ 0.075} \\
  & $\sigma$-GPT-random & 0.008 {\scriptsize $\pm$ 0.002} & 0.938 {\scriptsize $\pm$ 0.032} & 0.264 {\scriptsize $\pm$ 0.249} & 0.955 {\scriptsize $\pm$ 0.007} & 0.140 {\scriptsize $\pm$ 0.035} \\
  & MLM (BERT) & 0.132 {\scriptsize $\pm$ 0.070} & 0.918 {\scriptsize $\pm$ 0.019} & 0.820 {\scriptsize $\pm$ 0.078} & 0.928 {\scriptsize $\pm$ 0.055} & 0.552 {\scriptsize $\pm$ 0.125} \\
  & Diffusion & 0.065 {\scriptsize $\pm$ 0.013} & 0.894 {\scriptsize $\pm$ 0.023} & 0.800 {\scriptsize $\pm$ 0.022} & \textbf{0.958 {\scriptsize $\pm$ 0.009}} & 0.729 {\scriptsize $\pm$ 0.022} \\
  & GPT & \textbf{0.923 {\scriptsize $\pm$ 0.046}} & -- & -- & 0.919 {\scriptsize $\pm$ 0.030} & \textbf{0.878 {\scriptsize $\pm$ 0.063}} \\
\bottomrule
\end{tabular}%
}
\end{center}
\end{table*}

\begin{table*}[ht]
\caption{MAUVE Ablation (Base): MAUVE ($\uparrow$) on WikiText-103 across evaluation modes and conditioning scales using GPT-2 Base backbone.}
\label{tab:mauve_base}
\begin{center}
\resizebox{\columnwidth}{!}{%
\begin{tabular}{ll ccccc}
\toprule
\textbf{Scale} & \textbf{Model} & \textbf{Unconditional} & \textbf{Infilling} & \textbf{Training dist.} & \textbf{Infilling w/o future} & \textbf{Training dist. w/o future} \\
\midrule
\multirow{6}{*}{cond0-20}
  & AC-GPT (ours) & 0.887 {\scriptsize $\pm$ 0.052} & \textbf{0.964 {\scriptsize $\pm$ 0.003}} & \textbf{0.937 {\scriptsize $\pm$ 0.012}} & 0.933 {\scriptsize $\pm$ 0.014} & 0.887 {\scriptsize $\pm$ 0.066} \\
  & $\sigma$-GPT-temporal & \textbf{0.947 {\scriptsize $\pm$ 0.011}} & 0.935 {\scriptsize $\pm$ 0.004} & 0.936 {\scriptsize $\pm$ 0.048} & 0.926 {\scriptsize $\pm$ 0.036} & \textbf{0.892 {\scriptsize $\pm$ 0.027}} \\
  & $\sigma$-GPT-random & 0.012 {\scriptsize $\pm$ 0.005} & 0.790 {\scriptsize $\pm$ 0.117} & 0.012 {\scriptsize $\pm$ 0.004} & 0.646 {\scriptsize $\pm$ 0.076} & 0.016 {\scriptsize $\pm$ 0.007} \\
  & MLM (BERT) & 0.601 {\scriptsize $\pm$ 0.032} & 0.870 {\scriptsize $\pm$ 0.041} & 0.600 {\scriptsize $\pm$ 0.170} & 0.903 {\scriptsize $\pm$ 0.027} & 0.317 {\scriptsize $\pm$ 0.116} \\
  & Diffusion & 0.145 {\scriptsize $\pm$ 0.092} & 0.645 {\scriptsize $\pm$ 0.047} & 0.437 {\scriptsize $\pm$ 0.085} & \textbf{0.946 {\scriptsize $\pm$ 0.024}} & 0.646 {\scriptsize $\pm$ 0.191} \\
  & GPT & 0.932 {\scriptsize $\pm$ 0.043} & -- & -- & 0.929 {\scriptsize $\pm$ 0.019} & 0.849 {\scriptsize $\pm$ 0.040} \\
\cmidrule{1-7}
\multirow{6}{*}{cond0-40}
  & AC-GPT (ours) & \textbf{0.936 {\scriptsize $\pm$ 0.034}} & \textbf{0.940 {\scriptsize $\pm$ 0.016}} & 0.915 {\scriptsize $\pm$ 0.036} & \textbf{0.959 {\scriptsize $\pm$ 0.015}} & 0.847 {\scriptsize $\pm$ 0.059} \\
  & $\sigma$-GPT-temporal & 0.922 {\scriptsize $\pm$ 0.041} & 0.916 {\scriptsize $\pm$ 0.053} & \textbf{0.932 {\scriptsize $\pm$ 0.036}} & 0.954 {\scriptsize $\pm$ 0.022} & 0.783 {\scriptsize $\pm$ 0.051} \\
  & $\sigma$-GPT-random & 0.012 {\scriptsize $\pm$ 0.004} & 0.890 {\scriptsize $\pm$ 0.029} & 0.009 {\scriptsize $\pm$ 0.001} & 0.861 {\scriptsize $\pm$ 0.123} & 0.010 {\scriptsize $\pm$ 0.002} \\
  & MLM (BERT) & 0.629 {\scriptsize $\pm$ 0.059} & 0.939 {\scriptsize $\pm$ 0.014} & 0.806 {\scriptsize $\pm$ 0.025} & 0.926 {\scriptsize $\pm$ 0.033} & 0.431 {\scriptsize $\pm$ 0.124} \\
  & Diffusion & 0.178 {\scriptsize $\pm$ 0.032} & 0.828 {\scriptsize $\pm$ 0.020} & 0.571 {\scriptsize $\pm$ 0.124} & 0.908 {\scriptsize $\pm$ 0.066} & 0.482 {\scriptsize $\pm$ 0.064} \\
  & GPT & 0.932 {\scriptsize $\pm$ 0.043} & -- & -- & 0.922 {\scriptsize $\pm$ 0.016} & \textbf{0.862 {\scriptsize $\pm$ 0.004}} \\
\cmidrule{1-7}
\multirow{6}{*}{cond0-60}
  & AC-GPT (ours) & \textbf{0.941 {\scriptsize $\pm$ 0.020}} & 0.948 {\scriptsize $\pm$ 0.007} & \textbf{0.949 {\scriptsize $\pm$ 0.019}} & 0.937 {\scriptsize $\pm$ 0.008} & 0.739 {\scriptsize $\pm$ 0.097} \\
  & $\sigma$-GPT-temporal & 0.924 {\scriptsize $\pm$ 0.022} & \textbf{0.961 {\scriptsize $\pm$ 0.020}} & 0.941 {\scriptsize $\pm$ 0.023} & 0.937 {\scriptsize $\pm$ 0.021} & 0.734 {\scriptsize $\pm$ 0.100} \\
  & $\sigma$-GPT-random & 0.012 {\scriptsize $\pm$ 0.010} & 0.932 {\scriptsize $\pm$ 0.050} & 0.018 {\scriptsize $\pm$ 0.002} & 0.920 {\scriptsize $\pm$ 0.065} & 0.012 {\scriptsize $\pm$ 0.002} \\
  & MLM (BERT) & 0.812 {\scriptsize $\pm$ 0.062} & 0.934 {\scriptsize $\pm$ 0.024} & 0.832 {\scriptsize $\pm$ 0.015} & \textbf{0.945 {\scriptsize $\pm$ 0.032}} & 0.349 {\scriptsize $\pm$ 0.072} \\
  & Diffusion & 0.124 {\scriptsize $\pm$ 0.039} & 0.865 {\scriptsize $\pm$ 0.057} & 0.674 {\scriptsize $\pm$ 0.043} & 0.936 {\scriptsize $\pm$ 0.028} & 0.592 {\scriptsize $\pm$ 0.106} \\
  & GPT & 0.932 {\scriptsize $\pm$ 0.043} & -- & -- & 0.936 {\scriptsize $\pm$ 0.023} & \textbf{0.780 {\scriptsize $\pm$ 0.035}} \\
\cmidrule{1-7}
\multirow{6}{*}{cond0-80}
  & AC-GPT (ours) & 0.863 {\scriptsize $\pm$ 0.135} & 0.935 {\scriptsize $\pm$ 0.031} & \textbf{0.963 {\scriptsize $\pm$ 0.012}} & 0.948 {\scriptsize $\pm$ 0.014} & \textbf{0.852 {\scriptsize $\pm$ 0.053}} \\
  & $\sigma$-GPT-temporal & 0.828 {\scriptsize $\pm$ 0.094} & \textbf{0.971 {\scriptsize $\pm$ 0.004}} & 0.957 {\scriptsize $\pm$ 0.024} & 0.954 {\scriptsize $\pm$ 0.031} & 0.673 {\scriptsize $\pm$ 0.136} \\
  & $\sigma$-GPT-random & 0.011 {\scriptsize $\pm$ 0.003} & 0.947 {\scriptsize $\pm$ 0.033} & 0.075 {\scriptsize $\pm$ 0.098} & 0.952 {\scriptsize $\pm$ 0.030} & 0.037 {\scriptsize $\pm$ 0.017} \\
  & MLM (BERT) & 0.544 {\scriptsize $\pm$ 0.135} & 0.948 {\scriptsize $\pm$ 0.016} & 0.952 {\scriptsize $\pm$ 0.022} & 0.931 {\scriptsize $\pm$ 0.020} & 0.580 {\scriptsize $\pm$ 0.058} \\
  & Diffusion & 0.185 {\scriptsize $\pm$ 0.084} & 0.850 {\scriptsize $\pm$ 0.024} & 0.819 {\scriptsize $\pm$ 0.065} & 0.933 {\scriptsize $\pm$ 0.053} & 0.575 {\scriptsize $\pm$ 0.077} \\
  & GPT & \textbf{0.932 {\scriptsize $\pm$ 0.043}} & -- & -- & \textbf{0.959 {\scriptsize $\pm$ 0.013}} & 0.814 {\scriptsize $\pm$ 0.086} \\
\cmidrule{1-7}
\multirow{6}{*}{cond0-100}
  & AC-GPT (ours) & 0.892 {\scriptsize $\pm$ 0.059} & 0.950 {\scriptsize $\pm$ 0.025} & \textbf{0.954 {\scriptsize $\pm$ 0.026}} & 0.957 {\scriptsize $\pm$ 0.011} & 0.834 {\scriptsize $\pm$ 0.003} \\
  & $\sigma$-GPT-temporal & 0.897 {\scriptsize $\pm$ 0.058} & \textbf{0.954 {\scriptsize $\pm$ 0.040}} & 0.936 {\scriptsize $\pm$ 0.024} & \textbf{0.966 {\scriptsize $\pm$ 0.036}} & 0.753 {\scriptsize $\pm$ 0.112} \\
  & $\sigma$-GPT-random & 0.014 {\scriptsize $\pm$ 0.002} & 0.944 {\scriptsize $\pm$ 0.022} & 0.188 {\scriptsize $\pm$ 0.117} & 0.953 {\scriptsize $\pm$ 0.025} & 0.122 {\scriptsize $\pm$ 0.007} \\
  & MLM (BERT) & 0.651 {\scriptsize $\pm$ 0.041} & 0.926 {\scriptsize $\pm$ 0.004} & 0.934 {\scriptsize $\pm$ 0.050} & 0.914 {\scriptsize $\pm$ 0.046} & 0.591 {\scriptsize $\pm$ 0.120} \\
  & Diffusion & 0.062 {\scriptsize $\pm$ 0.007} & 0.900 {\scriptsize $\pm$ 0.044} & 0.819 {\scriptsize $\pm$ 0.053} & 0.964 {\scriptsize $\pm$ 0.009} & 0.802 {\scriptsize $\pm$ 0.105} \\
  & GPT & \textbf{0.932 {\scriptsize $\pm$ 0.043}} & -- & -- & 0.956 {\scriptsize $\pm$ 0.027} & \textbf{0.867 {\scriptsize $\pm$ 0.053}} \\
\bottomrule
\end{tabular}%
}
\end{center}
\end{table*}

\begin{table*}[ht]
\caption{MAUVE Ablation (Medium): MAUVE ($\uparrow$) on WikiText-103 across evaluation modes and conditioning scales using GPT-2 Medium backbone.}
\label{tab:mauve_medium}
\begin{center}
\resizebox{\columnwidth}{!}{%
\begin{tabular}{ll ccccc}
\toprule
\textbf{Scale} & \textbf{Model} & \textbf{Unconditional} & \textbf{Infilling} & \textbf{Training dist.} & \textbf{Infilling w/o future} & \textbf{Training dist. w/o future} \\
\midrule
\multirow{6}{*}{cond0-20}
  & AC-GPT (ours) & 0.922 {\scriptsize $\pm$ 0.047} & 0.948 {\scriptsize $\pm$ 0.014} & \textbf{0.955 {\scriptsize $\pm$ 0.008}} & 0.910 {\scriptsize $\pm$ 0.100} & 0.824 {\scriptsize $\pm$ 0.045} \\
  & $\sigma$-GPT-temporal & 0.937 {\scriptsize $\pm$ 0.029} & \textbf{0.949 {\scriptsize $\pm$ 0.036}} & 0.940 {\scriptsize $\pm$ 0.032} & \textbf{0.970 {\scriptsize $\pm$ 0.007}} & \textbf{0.872 {\scriptsize $\pm$ 0.084}} \\
  & $\sigma$-GPT-random & 0.015 {\scriptsize $\pm$ 0.007} & 0.764 {\scriptsize $\pm$ 0.148} & 0.018 {\scriptsize $\pm$ 0.008} & 0.786 {\scriptsize $\pm$ 0.132} & 0.013 {\scriptsize $\pm$ 0.004} \\
  & MLM (BERT) & 0.296 {\scriptsize $\pm$ 0.148} & 0.842 {\scriptsize $\pm$ 0.059} & 0.438 {\scriptsize $\pm$ 0.229} & 0.825 {\scriptsize $\pm$ 0.134} & 0.191 {\scriptsize $\pm$ 0.102} \\
  & Diffusion & 0.165 {\scriptsize $\pm$ 0.042} & 0.585 {\scriptsize $\pm$ 0.072} & 0.471 {\scriptsize $\pm$ 0.047} & 0.936 {\scriptsize $\pm$ 0.024} & 0.803 {\scriptsize $\pm$ 0.081} \\
  & GPT & \textbf{0.952 {\scriptsize $\pm$ 0.015}} & -- & -- & 0.946 {\scriptsize $\pm$ 0.036} & 0.863 {\scriptsize $\pm$ 0.062} \\
\cmidrule{1-7}
\multirow{6}{*}{cond0-40}
  & AC-GPT (ours) & 0.870 {\scriptsize $\pm$ 0.083} & 0.928 {\scriptsize $\pm$ 0.019} & 0.937 {\scriptsize $\pm$ 0.018} & 0.955 {\scriptsize $\pm$ 0.023} & 0.743 {\scriptsize $\pm$ 0.027} \\
  & $\sigma$-GPT-temporal & 0.918 {\scriptsize $\pm$ 0.019} & \textbf{0.937 {\scriptsize $\pm$ 0.025}} & \textbf{0.940 {\scriptsize $\pm$ 0.013}} & \textbf{0.978 {\scriptsize $\pm$ 0.006}} & 0.748 {\scriptsize $\pm$ 0.110} \\
  & $\sigma$-GPT-random & 0.010 {\scriptsize $\pm$ 0.005} & 0.805 {\scriptsize $\pm$ 0.080} & 0.007 {\scriptsize $\pm$ 0.000} & 0.854 {\scriptsize $\pm$ 0.044} & 0.012 {\scriptsize $\pm$ 0.003} \\
  & MLM (BERT) & 0.426 {\scriptsize $\pm$ 0.083} & 0.932 {\scriptsize $\pm$ 0.012} & 0.819 {\scriptsize $\pm$ 0.059} & 0.926 {\scriptsize $\pm$ 0.015} & 0.460 {\scriptsize $\pm$ 0.047} \\
  & Diffusion & 0.159 {\scriptsize $\pm$ 0.032} & 0.749 {\scriptsize $\pm$ 0.077} & 0.620 {\scriptsize $\pm$ 0.072} & 0.939 {\scriptsize $\pm$ 0.025} & 0.418 {\scriptsize $\pm$ 0.064} \\
  & GPT & \textbf{0.952 {\scriptsize $\pm$ 0.015}} & -- & -- & 0.926 {\scriptsize $\pm$ 0.020} & \textbf{0.832 {\scriptsize $\pm$ 0.041}} \\
\cmidrule{1-7}
\multirow{6}{*}{cond0-60}
  & AC-GPT (ours) & 0.936 {\scriptsize $\pm$ 0.035} & 0.940 {\scriptsize $\pm$ 0.040} & \textbf{0.946 {\scriptsize $\pm$ 0.008}} & \textbf{0.942 {\scriptsize $\pm$ 0.031}} & \textbf{0.786 {\scriptsize $\pm$ 0.169}} \\
  & $\sigma$-GPT-temporal & 0.931 {\scriptsize $\pm$ 0.035} & \textbf{0.960 {\scriptsize $\pm$ 0.019}} & 0.931 {\scriptsize $\pm$ 0.030} & 0.927 {\scriptsize $\pm$ 0.047} & 0.748 {\scriptsize $\pm$ 0.124} \\
  & $\sigma$-GPT-random & 0.019 {\scriptsize $\pm$ 0.018} & 0.947 {\scriptsize $\pm$ 0.021} & 0.228 {\scriptsize $\pm$ 0.378} & 0.887 {\scriptsize $\pm$ 0.029} & 0.015 {\scriptsize $\pm$ 0.009} \\
  & MLM (BERT) & 0.781 {\scriptsize $\pm$ 0.039} & 0.924 {\scriptsize $\pm$ 0.027} & 0.874 {\scriptsize $\pm$ 0.012} & 0.922 {\scriptsize $\pm$ 0.018} & 0.482 {\scriptsize $\pm$ 0.188} \\
  & Diffusion & 0.139 {\scriptsize $\pm$ 0.052} & 0.883 {\scriptsize $\pm$ 0.105} & 0.657 {\scriptsize $\pm$ 0.105} & 0.886 {\scriptsize $\pm$ 0.053} & 0.479 {\scriptsize $\pm$ 0.061} \\
  & GPT & \textbf{0.952 {\scriptsize $\pm$ 0.015}} & -- & -- & 0.940 {\scriptsize $\pm$ 0.018} & 0.709 {\scriptsize $\pm$ 0.125} \\
\cmidrule{1-7}
\multirow{6}{*}{cond0-80}
  & AC-GPT (ours) & 0.950 {\scriptsize $\pm$ 0.034} & 0.912 {\scriptsize $\pm$ 0.043} & 0.932 {\scriptsize $\pm$ 0.029} & \textbf{0.946 {\scriptsize $\pm$ 0.032}} & \textbf{0.847 {\scriptsize $\pm$ 0.049}} \\
  & $\sigma$-GPT-temporal & 0.817 {\scriptsize $\pm$ 0.056} & \textbf{0.943 {\scriptsize $\pm$ 0.028}} & \textbf{0.945 {\scriptsize $\pm$ 0.009}} & 0.942 {\scriptsize $\pm$ 0.037} & 0.842 {\scriptsize $\pm$ 0.084} \\
  & $\sigma$-GPT-random & 0.012 {\scriptsize $\pm$ 0.002} & 0.931 {\scriptsize $\pm$ 0.032} & 0.051 {\scriptsize $\pm$ 0.039} & 0.922 {\scriptsize $\pm$ 0.044} & 0.040 {\scriptsize $\pm$ 0.012} \\
  & MLM (BERT) & 0.568 {\scriptsize $\pm$ 0.149} & 0.931 {\scriptsize $\pm$ 0.030} & 0.854 {\scriptsize $\pm$ 0.080} & 0.932 {\scriptsize $\pm$ 0.034} & 0.537 {\scriptsize $\pm$ 0.026} \\
  & Diffusion & 0.130 {\scriptsize $\pm$ 0.027} & 0.870 {\scriptsize $\pm$ 0.023} & 0.813 {\scriptsize $\pm$ 0.064} & 0.915 {\scriptsize $\pm$ 0.013} & 0.616 {\scriptsize $\pm$ 0.144} \\
  & GPT & \textbf{0.952 {\scriptsize $\pm$ 0.015}} & -- & -- & 0.927 {\scriptsize $\pm$ 0.008} & 0.805 {\scriptsize $\pm$ 0.077} \\
\cmidrule{1-7}
\multirow{6}{*}{cond0-100}
  & AC-GPT (ours) & 0.887 {\scriptsize $\pm$ 0.039} & \textbf{0.952 {\scriptsize $\pm$ 0.018}} & \textbf{0.958 {\scriptsize $\pm$ 0.028}} & \textbf{0.967 {\scriptsize $\pm$ 0.009}} & \textbf{0.884 {\scriptsize $\pm$ 0.067}} \\
  & $\sigma$-GPT-temporal & 0.933 {\scriptsize $\pm$ 0.037} & 0.950 {\scriptsize $\pm$ 0.029} & 0.942 {\scriptsize $\pm$ 0.032} & 0.963 {\scriptsize $\pm$ 0.030} & 0.831 {\scriptsize $\pm$ 0.051} \\
  & $\sigma$-GPT-random & 0.014 {\scriptsize $\pm$ 0.005} & 0.932 {\scriptsize $\pm$ 0.012} & 0.094 {\scriptsize $\pm$ 0.012} & 0.889 {\scriptsize $\pm$ 0.044} & 0.098 {\scriptsize $\pm$ 0.021} \\
  & MLM (BERT) & 0.647 {\scriptsize $\pm$ 0.014} & 0.943 {\scriptsize $\pm$ 0.039} & 0.945 {\scriptsize $\pm$ 0.024} & 0.953 {\scriptsize $\pm$ 0.033} & 0.681 {\scriptsize $\pm$ 0.085} \\
  & Diffusion & 0.178 {\scriptsize $\pm$ 0.080} & 0.936 {\scriptsize $\pm$ 0.009} & 0.803 {\scriptsize $\pm$ 0.030} & 0.957 {\scriptsize $\pm$ 0.014} & 0.807 {\scriptsize $\pm$ 0.062} \\
  & GPT & \textbf{0.952 {\scriptsize $\pm$ 0.015}} & -- & -- & 0.956 {\scriptsize $\pm$ 0.012} & 0.860 {\scriptsize $\pm$ 0.032} \\
\bottomrule
\end{tabular}%
}
\end{center}
\end{table*}

%% file: generation_samples.tex
\subsection{Sampling examples}\label{app:gen-samples}

To complement the aggregate MAUVE scores, we show qualitative samples for a single WikiText-103 article. For \emph{Unconditional generation} each method is given a 20-token prompt and generates the continuation freely; we show the causal GPT baseline, AC-GPT, $\sigma$-GPT-temporal, and $\sigma$-GPT-random. For \emph{Infilling} and \emph{Conditional generation} (training-distribution-like; $r_{\max} = 0.4$) we show the three any-order methods (AC-GPT, $\sigma$-GPT-temporal, and $\sigma$-GPT-random); the causal baseline is omitted because these modes condition on future context, which it cannot perform.

\subsubsection{Unconditional generation}\label{app:gen-samples:uncond}

\begin{tcolorbox}[breakable,title={GPT (baseline)}]
\footnotesize \hl{ = Homarus gammarus =

 Homarus gammarus , known as the European lobster} , is a species of species of the family Bovidae , found in the Mediterranean Sea . It was described by Edward Drinker Cope in 1776 as a new species of Homarus , but is currently unknown . It is currently recognized as a separate species , Homarus v. cyathus . It is similar to the American lobster , but has a shorter body and a longer body than the French lobster ( Pharus fuscatus ) . The common name is derived from the Latin word " sickle " , and the specific epithet v. cyathus is the Latin word for " sickle " .

@ running , it is found in the Mediterranean Sea , and is found in the Mediterranean Sea . It has been described as a separate species from the British lobster ( Pharus fuscatus ) .

ive measures of its body size have been used to estimate the number of individuals of this species . A female collected off Spain in 1776 was said to be the largest individual of the species , weighing 4.8 kg ( 9.5 lb ) . The female is approximately the same size as the male , with a body length of 15 to 20 cm ( 5.9 to 7.9 in ) . The female is about as large as the male , and has a body length of 11 to 12 cm ( 4.3 to 4.7 in ) . The male has a body length of 17 to 20 cm ( 6.7 to 7.9 in ) , while the female is much smaller than the male . It has a body length of 11 to 13 cm ( 4.3 to 5.1 in ) .

@ running and mating are thought to occur in the spring or autumn . The eggs of Homarus v. cyathus hatch in the spring and summer , and the young are born in spring . The juveniles are not sexually mature until they are five to ten months old . The hatchlings are light brown to reddish brown in color , with a white underside . They are also very similar to the American lobster . The European lobster has a short , fairly long body , but the head is larger than the body . It has a very long body and long arms . The female has a large head and neck , and a large head with a broad , flattened head . The juvenile has a head that is shorter than the adult , and a shorter body . The juvenile has a body length of up to 15 cm ( 5.9 in ) .

bourg in France stated that this species can live up to 20 years .

@ to-year studies of the stomach contents of this species have shown that the stomach contents of Homarus v. cyathus are similar to those of the French lobster , but are different from those of the French lobster . A study of the stomach contents of the French lobster found that Homarus v. cyathus was also similar to the French lobster , but larger , with a larger body size and a longer body .

@ to-life feeding is typical of the French species . The French lobster is a small , robust , and long-tailed species , with the head and body length of the French lobster as much as 1.5 to 2 cm ( 0.59 to 0.79 in ) . The fins of the French lobster are hooked .

@ running and mating is thought to occur in spring or autumn , with females laying two to three eggs .

 paint is used to protect the eggs from predation .

@ running larvae are usually found in the spring , but may also be found in the autumn . In the Mediterranean Sea , the larvae are known as " breakers " , because they are intermediate between the adults and the adults . They are usually found in the Mediterranean Sea and the Mediterranean Sea , but may also be found in the Black Sea , the Black Sea , and the Black Sea .

 view of the two species is based on the very different methods used to identify and distinguish the two . The French species has a body length of 1.5 to 3.5 cm ( 0.71 to 1.18 in ) , a body length of 2.5 to 5.5 cm ( 1.0 to 2.0 in ) , and a body mass of 0.4 to 1.7 g ( 0.067 to 0.071 oz ) . The French species is also known from the Black Sea and the Black Sea , but the French and Italian species differ in the details of the body and body .

 collectively , the French species is the smallest of the French species . The head is longer than the body , and the head and body length
\end{tcolorbox}

\begin{tcolorbox}[breakable,title={AC-GPT (ours)}]
\footnotesize \hl{ = Homarus gammarus =

 Homarus gammarus , known as the European lobster} gammarus , is a species of round-shaded fish native to the Atlantic Ocean , including the Caribbean Sea , and the Mediterranean Sea . It is a relatively large species , measuring 7.5 -- 8.5 cm ( 3.0 -- 3.1 in ) long . It is a bottom-dwelling species , and feeds mainly on polychaete worms . It is a bottom-dwelling species , typically 6 -- 15 cm ( 2.4 -- 5.9 in ) long , with a maximum known length of 1.6 -- 2.5 m ( 4.9 -- 6.2 ft ) . It is the only member of its genus to have a single functional ovary and two functional uteruses .

D-D. formosa is a fairly common species , although it is rarely encountered . It is fairly common and feeds on polychaete worms and small benthic bony fishes , and has been found in coastal waters . It is aplacental viviparous , with the developing embryos sustained to term by a placental connection to the mother . Females give birth to up to six pups every other year . This species is aplacental viviparous , with the developing embryos sustained to term by yolk . Females bear litters of two to four pups every other year . Mating occurs from May to July , with the embryos biting the bottom for each other , and the developing embryos are placental viviparous . The embryos are sustained to term by a placental connection to the mother . The International Union for Conservation of Nature ( IUCN ) has assessed this species as Near Threatened .

 Number of requiem sharks worldwide , it is one of the most abundant sharks in the Atlantic . This species is viviparous , with the developing embryos sustained to term by a placental connection to the mother . Females bear litters of one to 12 pups every other year , following a gestation period of four months . The young are born in large numbers , and are capable of changing their mother 's coloration . The International Union for Conservation of Nature ( IUCN ) has assessed this species as Vulnerable , as it has a large population and is of " least concern " .

field data indicate that the International Union for Conservation of Nature ( IUCN ) has assessed this species as Near Threatened .

away and slow-moving , the slow-moving , sickle-shaped pups grow larger than adults and are thus less susceptible to population depletion . There is a recent increase in the number of sharks in the diet of the slow-moving , pelagic longline fisheries in the Mediterranean Sea , and the number of rays caught off the coasts of North Africa and the Mediterranean . This species is not regarded as a large species , and is therefore classed as Vulnerable by the International Union for Conservation of Nature ( IUCN ) .

ming , or " shark finning " , is the most common form of shark fishing in the Mediterranean , with a total catch of 2.5 million tonnes ( 2.3 million short tons ) in 2007 . The shark is caught incidentally in bottom trawls and bottom trawls , and sold fresh or dried . The International Union for Conservation of Nature ( IUCN ) has assessed this species as Near Threatened , as it faces a threat of extinction due to its large numbers and large numbers , but this species is not considered to be of least concern .

 Images of the Mediterranean Sea and Canary Islands are used by commercial fisheries targeting this species . A number of commercial fisheries targeting this species have been reported in the Mediterranean , including in the North Sea , the Mediterranean , and off the Azores . The number of sharks caught off Portugal and Spain has not been quantified , but the numbers of sharks caught off Portugal are small and poorly documented . The International Union for Conservation of Nature ( IUCN ) has assessed this species as Near Threatened , as it is not found in any protected areas , and has been assessed as Vulnerable .

way fisheries targeting this species include the Mediterranean and Atlantic , with the exception of the Atlantic and Mediterranean . The meat is sold fresh or dried , with the fins taken off Portugal and the United States . It is also caught in lobster traps off the Bahamas , and off the Bahamas . The meat is sold fresh or dried , or processed into fishmeal . The meat is sold fresh or dried and salted , and is sold in small quantities in markets in the United States and the UK . In North America , the meat is sold fresh or dried , and is sold fresh or dried . The US Navy has instituted strict fishery regulation , and has set up a fishery management
\end{tcolorbox}

\begin{tcolorbox}[breakable,title={$\sigma$-GPT-temporal}]
\footnotesize \hl{ = Homarus gammarus =

 Homarus gammarus , known as the European lobster} , is a species of lobster and lobster in the family Proteaceae . This species is found in Europe , North America , and the Caribbean Sea , and is typically found in tropical and subtropical waters from the Gulf of Mexico to the Gulf of Mexico . The species is a large lobster , but is sometimes larger than it is wide . The head is large and pointed , with a long , narrow head , and a broad mouth . The upperparts are generally darker , with a brownish blotch on the head . The body is light grayish to brownish , with an irregular pattern of darker blotches on the sides and a lighter blotch on the sides . The mouth is wide and has a deep groove on the upper jaw , and a deep groove on the upper jaw . The two species are distinguished by differences in the colouration of their gills , which are black with a white margin , and a narrow , circular lobed pectoral fin . The species is smaller than the other members of its genus , and the distribution of this species is unknown .

@ B. gammarus is a slow-growing lobster , and has been recorded from southern Spain , the Canary Islands , and the Canary Islands . It is found in the Atlantic Ocean , from southern Spain to Spain , and has been recorded from the Canary Islands , the Azores , and Madeira . It is a bottom-dwelling species with a strong head and a long , broad head . The diet of this species is mainly nocturnal , and its diet includes cephalopods , worms , and other crustaceans .

@ B. gammarus is a slow-growing lobster , and its maximum recorded weight is 135 kg ( 298 lb ) . The maximum recorded weight is 12 kg ( 26 lb ) . The diet of this species is varied , but it is most active at night , and is mainly active during the day . Females are ovoviviparous , with the developing embryos nourished by yolk . The developing embryos are pigmented with an oily substance . The eggs are white , smooth , and later maturing into small , oval , obtuse mouths with a row of small , circular pores . The eggs are brownish-brown to dark brown , and are covered with a fine layer of mucus . The eggs are white , with a brownish or greyish blotch on the sides and a slightly darker blotch on the sides . The eggs are oval to oval , with a broadly elliptical shape . The egg is laid at a shallow , smooth rate , with a slight indentation in the middle . The eggs are small , measuring 14 -- 16 by 7 -- 9 ?m . The eggs are incubated by the female for 17 -- 18 days , and the male for 33 -- 39 days . The eggs are dark brownish-brown and have a pale blotch with a deep blotch on the sides . The eggs are white , smooth , and measure 20 -- 30 by 4.5 -- 6 ?m . The eggs are brownish-brown and measure 7 -- 15 by 4 -- 6 ?m . The eggs are light brownish-brown with a slight brown blotch on the upper surface , and are covered with brownish granules .

@ B. gammarus is one of several species in the genus Proteus to which the true species name is given . It is the only species in the genus Proteus to have a true male .

@ B. gammarus has been recorded from the Canary Islands . The type locality is in the Canary Islands , where it was found in a bay with a population of 13 , and there are records of sightings in the area .

@ B. gammarus has been reported from the Canary Islands , and from the Canary Islands .

 Dolichthys sanguineus , a species of Pseudoryzomys , was reported from a bay in the Canary Islands in 1984 . It was found on the Canary Islands , and was found to be common in the Gulf of Mexico .

 Ginura sanguineus , a species of Pseudoryzomys , is found in the Canary Islands and the Canary Islands .

 NLN-Epp . , a species of Pseudoryzomys , was recorded from the Canary Islands and Madeira in 1978 . It was found in the Canary Islands , and from the Canary Islands .

@ B. sanguineus is also found in the Canary Islands .

 Purpurea sanguineus , a species of Pseudoryzomys , is found in the Canary Islands , and from the Canary Islands .

 painting shows a single male specimen , which was previously thought to be a female .

 11-to-12-year
\end{tcolorbox}

\begin{tcolorbox}[breakable,title={$\sigma$-GPT-random}]
\footnotesize \hl{ = Homarus gammarus =

 Homarus gammarus , known as the European lobster} , is a species of many fish , caught in the Mediterranean and off the southern Atlantic . It is known from the family Mycogidae , the family Polycena , which is commonly known used as " oph " " , a Latin for the Latin word " , " the Latin word ) . Its name name is derived from the Latin word of the cromosos , meaning " black " " and " ' ophos " " . Its name is derived from the Latin words kocososososososososo\discretionary{}{}{}sosososososososososo\discretionary{}{}{}sosososososososososo\discretionary{}{}{}sosososososososososo\discretionary{}{}{}sosososososososososo\discretionary{}{}{}sosososososososososo\discretionary{}{}{}sosososososososososo\discretionary{}{}{}sosososososososososo\discretionary{}{}{}sosososososososososo\discretionary{}{}{}sosososososososososo\discretionary{}{}{}sosososososososososo\discretionary{}{}{}sosososososososososo\discretionary{}{}{}sosososososososososo\discretionary{}{}{}sosososososososososo\discretionary{}{}{}sosososososososososo\discretionary{}{}{}sosososososososososo\discretionary{}{}{}sosososososososososo\discretionary{}{}{}sosososososososososo\discretionary{}{}{}sosososososososososo\discretionary{}{}{}sosososososososososo\discretionary{}{}{}sosososos ,osos ofos ,osososososososososos , ,osososososososososos inos @ ,osos ) ) and ) and and ,os theos Theos .os (os "os ,osos .os .os , = of ands a " "os ofos and .os of ,ososososososos andosasosososososososos and the cerosos The "os a mosos the plos , Theos .os foros ofos @oss ofos ,osos ,osos andos

 theos ,osos of aos inos ofos from ,os .osasos ofos the dosos "os toos of .os that andos .os .os inos Theos , aos of .os , ofos of aos theos )os theos @ theos of .  asos .os in and , and and and and ,.os . ,  of theos of , . theos in " asos of @ theos , , , , 'osos and . as . .-os . , of . , , to the and  , aos ofos )os of @ oros . , of or theos , is , . . in of . , . theos of ofos onos ( theos "os " and , the " and the , " . the foros theos @ the theos "os . Theos . theos . . @ ,osos @ "os , of . = in . , , , theososos = to of of ,@osos , @ =os . . =osos ,- andososososos , of =os ( ,os "os

os ,ososos , ) =osotosos =osososos to haveos , theos ,os andosososos , )osososososos , and and of . =os =osos . theos of , Theos . os .os , .os of , , osos .osososos , theos and . = in . = . theos = . and , aos . . . the in is =os@ = =os ofos . , = )

osos as = , and . . , @ " =os , . the are =os . the = to . theos , . . and to of , and ,@ with andos the ofos , theos in . the Theos a , in " , . , . . , the , . as , the . , the are . . of . the .@s os in of

os and =os .os andos of , the , the of . . "os , The the , of " . . . . . in , to ,os

 = = on the with The@  .  of , . . the "os and and . a . . and the and the . . as , the = of , . . , the . the

 , The os the = =os for in of . . : a the and , . in =ossos in )os , , The- . of "os by to " a of @os a = os , osos theos
\end{tcolorbox}

\subsubsection{Infilling}\label{app:gen-samples:infill}

\begin{tcolorbox}[breakable,title={Source passage (\hl{shaded} tokens given to the model)}]
\footnotesize \hl{ = Homarus gammarus = }

\hl{ Homarus gammarus , known as the European lobster or common lobster , is a species of clawed lobster from the eastern Atlantic Ocean , Mediterranean Sea and parts of the Black Sea . It is closely related to the American lobster , H.} americanus . It may grow to a length of 60 cm ( 24 in ) and a mass of 6 kilograms ( 13 lb ) , and bears a conspicuous pair of claws . In life , the lobsters are blue , only becoming " lobster red " on cooking . Mating occurs in the summer , producing eggs which are carried by the females for up to a year before hatching into planktonic larvae . Homarus gammarus is a highly esteemed food , and is widely caught using lobster pots , mostly around the British Isles .

 = = Description = =

 Homarus gammarus is a large crustacean , with a body length up to 60 centimetres ( 24 in ) and weighing up to 5 -- 6 kilograms ( 11 -- 13 lb ) , although the lobsters caught in lobster pots are usually 23 -- 38 cm ( 9 -- 15 in ) long and weigh 0.7 -- 2.2 kg ( 1.5 -- 4.9 lb ) . Like other crustaceans , lobsters have a hard exoskeleton which they must shed in order to grow , in a process called ecdysis ( moulting ) . This may occur several times a year for young lobsters , but decreases to once every 1 -- 2 years for larger animals .

 The first pair of pereiopods is armed with a large , asymmetrical pair of claws . The larger one is the " crusher " , and has rounded nodules used for crushing prey ; the other is the " cutter " , which has sharp inner edges , and is used for holding or tearing the prey . Usually , the left claw is the crusher , and the right is the cutter .

 The exoskeleton is generally blue above , with spots that coalesce , and yellow below . The red colour associated with lobsters only appears after cooking . This occurs because , in life , the red pigment astaxanthin is bound to a protein complex , but the complex is broken up by the heat of cooking , releasing the red pigment .

 The closest relative of H. gammarus is the American lobster , Homarus americanus . The two species are very similar , and can be crossed artificially , although hybrids are unlikely to occur in the wild since their ranges do not overlap . The two species can be distinguished by a number of characteristics :

 The rostrum of H. americanus bears one or more spines on the underside , which are lacking in H. gammarus .

 The spines on the claws of H. americanus are red or red-tipped , while those of H. gammarus are white or white-tipped .

 The underside of the claw of H. americanus is orange or red , while that of H. gammarus is creamy white or very pale red .

 = = Life cycle = =

 Female H. gammarus reach sexual maturity when they have grown to a carapace length of 80 -- 85 millimetres ( 3.1 -- 3.3 in ) , whereas males mature at a slightly smaller size . Mating typically occurs in summer between a recently moulted female , whose shell is therefore soft , and a hard-shelled male . The female carries the eggs for up to 12 months , depending on the temperature , attached to her pleopods . Females carrying eggs are said to be " berried " and can be found throughout the year .

 The eggs hatch at night , and the larvae swim to the water surface where they drift with the ocean currents , preying on zooplankton . This stage involves three moults and lasts for 15\hl{ -- 35 days . After the third moult , the juvenile takes on a form closer to the adult , and adopts a benthic lifestyle . The juveniles are rarely seen in the wild , and are poorly known , although they are known to be capable of digging extensive burrows . It is estimated that only 1 larva in every 20,000 survives to the benthic phase . When they reach a carapace length of 15 mm ( 0.59 in ) , the juveniles leave their burrows and start their adult lives . }

\hl{ = = Distribution = = }

\hl{ Homarus gammarus is found across the north-eastern Atlantic Ocean from northern Norway to the Azores and Morocco , not including the Baltic Sea . It is also present in most of the Mediterranean Sea , only missing from the section east of Crete , and along only the north-west coast of the Black Sea . The northernmost populations are found in the Norwegian fjords Tys}f
\end{tcolorbox}

\begin{tcolorbox}[breakable,title={$\sigma$-GPT-temporal}]
\footnotesize  = Homarus gammarus =

 Homarus gammarus , known as the European lobster or common lobster , is a species of clawed lobster from the eastern Atlantic Ocean , Mediterranean Sea and parts of the Black Sea . It is closely related to the American lobster , Hromus californianus , and the black lobster , H. cuneatus , which are also similar in appearance . It is found from the western Atlantic Ocean to the Azores and Madeira . It is an important parasite of the Atlantic puffball ( H. helvipinna ) . It is a large , slender lobster with a robust body and a long , thick , cylindrical head . It is aplacental viviparous , meaning the developing embryos are nourished by yolk . Females bear two to seven litters of one to seven young every other year .

@ Pleistocene deposits were found in Iceland , and the oldest specimen ever found in the United States . The species is common and widespread , with numbers increasing in the western Atlantic and Atlantic , with populations increasing in the Gulf of Mexico , the Gulf of Mexico , and the Caribbean .

@

@ ? ?p?r?r / or ?p?r?r / is a species of lobster , and is closely related to the American lobster . It has a distinctive spore shape , which is unusual for lobster in having a long , slender stem . In the Atlantic , it is found from North Carolina to Brazil , and off West Africa .

@  ? ?r / is the most common lobster species in the Atlantic , with a worldwide distribution as far north as Mexico and as far south as South America . It is characterised by a robust body , a thin body , and a large head . Its two large dorsal fins are small and rounded , with a pointed apex . The anal fin is short and rounded , and the pectoral fins are very large . The underside is white , with a grayish-brown or grayish tint to the fins . The body is bilaterally symmetrical , with the anterior margin almost equal to the pectoral fin rear .

@  ? ?r / is found from southern France and Spain through the British Isles to Greenland and the eastern British Isles , and from Madeira , south to Mexico , the Caribbean , and the Caribbean . It has been introduced to the Caribbean and parts of the Caribbean . The Atlantic puffball has a circumpolar distribution in tropical and subtropical waters , and migrates from the Atlantic and Pacific Oceans . It is a non-migratory species , and has a circumpolar distribution , from the northern Atlantic to South America , including the coasts of the Azores and Madeira , to the Atlantic coast of the Azores and the Canary Islands .

@ ?r ?r ?r / is the only Atlantic lobster species from which the Atlantic puffball is native . The species has a worldwide distribution , but is most common in the western Atlantic Ocean , including the Mediterranean Sea , the Azores , and the North Atlantic , and also in the Mediterranean Sea , the Black Sea , the Azores , and the Black Sea . The Atlantic puffball is one of the few to inhabit the open ocean , where it is a bottom predator , and the Caribbean populations are migratory . The Atlantic puffball is a bottom predator with a varied prey range , though it is most commonly found near the surface .

@ ?r ?r ?r ?r ?r ?r is a polychaete predator that tends to feed on polychaete worms , crustaceans , and molluscs . The Atlantic puffball is an opportunistic predator , but will scavenge off soft materials , from burrows , and from objects such as the leathery crustaceans ( 15 -- 35 days . After the third moult , the juvenile takes on a form closer to the adult , and adopts a benthic lifestyle . The juveniles are rarely seen in the wild , and are poorly known , although they are known to be capable of digging extensive burrows . It is estimated that only 1 larva in every 20,000 survives to the benthic phase . When they reach a carapace length of 15 mm ( 0.59 in ) , the juveniles leave their burrows and start their adult lives .

 = = Distribution = =

 Homarus gammarus is found across the north-eastern Atlantic Ocean from northern Norway to the Azores and Morocco , not including the Baltic Sea . It is also present in most of the Mediterranean Sea , only missing from the section east of Crete , and along only the north-west coast of the Black Sea . The northernmost populations are found in the Norwegian fjords T
\end{tcolorbox}

\begin{tcolorbox}[breakable,title={AC-GPT (ours)}]
\footnotesize  = Homarus gammarus =

 Homarus gammarus , known as the European lobster or common lobster , is a species of clawed lobster from the eastern Atlantic Ocean , Mediterranean Sea and parts of the Black Sea . It is closely related to the American lobster , H. calendula . It is most closely related to H. calendula , which is the most similar species and is found on the Mediterranean coast of Europe . The two species are known to breed on sand . It is usually found growing on sandy flats , especially on the coasts of the Mediterranean and in the Mediterranean Sea , but can also be found in the Mediterranean .

 bycatch of deep sea fisheries on longlines in the Mediterranean and in the Black Sea . It is often caught in gill nets or on land , and caught as a pest in lobster traps . In the Mediterranean , it is caught as a bycatch of commercial trawls . In the Mediterranean , it is caught as a bycatch of longline commercial trawls , and in tuna trawls . It is also caught by trawl fisheries as a bycatch of trawls . Like other members of its family , it is viviparous with oophagous embryos containing male and female reproductive organs . The International Union for Conservation of Nature ( IUCN ) has assessed this species as Near Threatened .

 Sheridan 's original description of the Mediterranean lobster was based on a description by Henri Marie Ducrotay de Blainville in 1771 . In 1879 , British entomologist John Edward Gray included it in his 1810 work account of a lobster caught in the Black Sea . The genus Homarus is placed in its own monotypic genus Homarus , along with H. calendula and H. calendula . Molecular phylogenetic studies based on nuclear and mitochondrial DNA suggest Homarus may be a sister species to H. calendula , H. calendula , H. calendula , H. calendula , H. calendula and H. calendula . Genetic evidence suggests that Homarus forms a sister group to H. calendula , H. calendula and H. calendula .

 indicates that this species is distinct from H. calendula and H. calendula in the pattern of external coloration . It is a little larger than H. calendula , and the only member of its genus that is more similar to H. calendula . Its larger size and tarseness also suggests that this species may be the same size as H. calendula . The European lobster is less closely related to H. calendula than to H. calendula , and there are discrepancies in the geographical distribution of the three species .

DUS sequencing of internal DNA of the Mediterranean lobster and the American lobster were analysed using mitochondrial DNA to separate the two species . The authors suggested that the northern Mediterranean populations are genetically closer to the North American populations , and also that the two species are genetically closer to each other . The taxonomic history of the Mediterranean lobster is known from two morphological analyses of DNA sequences . In these analysis , the northern Mediterranean populations are genetically closer to H. calendula and H. calendula , and the southern Mediterranean populations are genetically closer to H. calendula than to H. calendula .

ball morphology is the most consistent among the two species . It is also a member of the H. calendula complex , which also includes the brown lobster , H. calendula and H. calendula . It is also a member of the H. calendula complex , along with H. calendula and H. calendula , H. calendula , H. calendula and H. calendula .

 Quarterback and Cancerodon species are closely related .

373

 overhaul

? --- ZN

ley \& Gladden

 responds in a laboratory . Results 15 -- 35 days . After the third moult , the juvenile takes on a form closer to the adult , and adopts a benthic lifestyle . The juveniles are rarely seen in the wild , and are poorly known , although they are known to be capable of digging extensive burrows . It is estimated that only 1 larva in every 20,000 survives to the benthic phase . When they reach a carapace length of 15 mm ( 0.59 in ) , the juveniles leave their burrows and start their adult lives .

 = = Distribution = =

 Homarus gammarus is found across the north-eastern Atlantic Ocean from northern Norway to the Azores and Morocco , not including the Baltic Sea . It is also present in most of the Mediterranean Sea , only missing from the section east of Crete , and along only the north-west coast of the Black Sea . The northernmost populations are found in the Norwegian fjords Tys
\end{tcolorbox}

\begin{tcolorbox}[breakable,title={$\sigma$-GPT-random}]
\footnotesize  = Homarus gammarus =

 Homarus gammarus , known as the European lobster or common lobster , is a species of clawed lobster from the eastern Atlantic Ocean , Mediterranean Sea and parts of the Black Sea . It is closely related to the American lobster , H the T. b. helicola , and the Latin epithet , " karu " , which translates in Latin for a " of " white " , which is derived from the Greek haiokos . The species is a derived derived species of the ceposus " , and is derived a genus of " lizard " , meaning the Ancient "-foot " " , and the head " , and and " head and the and " foot " " . The . The Latin = = = = Life = = = = = =   = =  = = =   S The = =  =  =

 =

 The Thearus Thearus =arus Thearus thearus p T thearustharus and ,

 Ar C thearus ( Myarus thearus  ( =. at S T Sarus My arusidae

 ich ich Unicorn

cn Pro Platterusus ( = = = = =  =

 = = = = =  =

 = = = =  = = = = = = = = = = = = = =  = = = = = = = = =  =  = = = = = = = = = = = = = = = = = = =  = = =  = = =  = = = = = Ext = = = = = =   = = = = = Predators  =

 = = = = = = = = = =  = = = = = = = = = = = = = = = = = = = = = = = = = = = = = = = = =   = = = = = = = = = =  =  = = = = = = = = = = = = = = = = = =  =  = = = = = = = = = = = = = = = = =  = = = = =  = =  = = = = = = = =   = = = = = = = = = = =  = = = = =  = = = = = = = = = = = = = = = = = = = = = = = = = = = = = = = = =  =  = = = =  = = = = = = =  = = = = = = = = = = = = = = = =  = = = = =  = =   = = = = = = = = = = = = = = = =     = = = = = = = = = = = = = = = = = = = = = = = = = = = = = = = = = = = = = =  = = = = = = = = = = = = = = = = = =  =    = = = = = = = = = = = = =   = = = = = = = = = = = = =  = = =  = = = =  = = = = = = = = = = = = = = = = = = = = = = = = = = = = = = = = = = = = = = =  = = = = = = = = = = = = = = = = = = = = = = = = = = =  = = = = = = = = = =  = = = = = = = = = = = = =  = = = = =  = = = = = = = = = = = = =

 = = = = Breedingeding = = = = = =

 Breedinging is 15 -- 35 days . After the third moult , the juvenile takes on a form closer to the adult , and adopts a benthic lifestyle . The juveniles are rarely seen in the wild , and are poorly known , although they are known to be capable of digging extensive burrows . It is estimated that only 1 larva in every 20,000 survives to the benthic phase . When they reach a carapace length of 15 mm ( 0.59 in ) , the juveniles leave their burrows and start their adult lives .

 = = Distribution = =

 Homarus gammarus is found across the north-eastern Atlantic Ocean from northern Norway to the Azores and Morocco , not including the Baltic Sea . It is also present in most of the Mediterranean Sea , only missing from the section east of Crete , and along only the north-west coast of the Black Sea . The northernmost populations are found in the Norwegian fjords T
\end{tcolorbox}

\subsubsection{Conditional generation}\label{app:gen-samples:cond}

\begin{tcolorbox}[breakable,title={Source passage (\hl{shaded} tokens given to the model)}]
\footnotesize \hl{ =} Homarus gammarus =

\hl{ Hom}arus gammarus , known as the European lobster or common lobster\hl{ , is a} species of clawed lobster from the eastern Atlantic Ocean\hl{ , Mediterranean Sea} and parts of the Black Sea . It is closely related to the American lobster , H. americanus . It may grow to a length\hl{ of 60} cm ( 24 in\hl{ )} and a mass\hl{ of 6 kilograms} (\hl{ 13} lb ) , and bears a conspicuous\hl{ pair of claws .} In life , the lobsters\hl{ are blue} , only becoming " lobster red " on cooking . Mating occurs in the summer , producing eggs which are carried by the females for up to a year before hatching into plank\hl{tonic} larvae . Homarus\hl{ gammar}us is\hl{ a highly} esteemed food , and is widely caught using lobster pots , mostly around the British Isles .

 = = Description = =

 Homarus gammarus is a large crustacean\hl{ ,} with a body length up to 60 centimetres ( 24 in ) and weighing up to 5 -- 6 kilograms ( 11 -- 13 lb ) , although the lobsters caught in\hl{ lobster} pots are usually 23 -- 38 cm ( 9 --\hl{ 15 in )} long and weigh 0.7\hl{ --} 2\hl{.2} kg ( 1\hl{.5 -- 4}.9 lb ) .\hl{ Like other} crustace\hl{ans ,} lobsters have a hard ex\hl{oskeleton which} they must shed in order to grow , in a process called ecdysis ( m\hl{oulting} ) . This may occur several times a year for young lobsters\hl{ , but decreases} to once every 1 -- 2 years for larger animals .

 The first pair of pereiopods is armed with a large , asymmetrical pair of claws .\hl{ The larger} one is the " crusher " , and has rounded nodules used for crushing prey ;\hl{ the} other is the " cutter " ,\hl{ which has sharp} inner edges , and is used for holding or tearing the prey . Usually , the left claw is the crusher , and the right is the cutter .

 The exos\hl{keleton is generally} blue above , with spots that coalesce , and yellow below . The red colour associated with lobsters only appears after cooking .\hl{ This occurs because} , in life , the red pigment astaxanthin is bound to a protein complex , but the complex is broken up\hl{ by the} heat of cooking , releasing the red pigment .

 The closest relative\hl{ of H.} gammarus is the American lobster , Homarus americanus .\hl{ The two} species are very similar , and can be crossed artificially , although hybrids are unlikely to occur in the wild since their ranges\hl{ do not} overlap . The two species can be distinguished by a\hl{ number of} characteristics :

 The rostrum of H. americanus bears one or more spines on the underside , which are lacking in H\hl{. gammarus} .

 The spines on the claws of H. americ\hl{anus are} red or red-tipped , while those of H\hl{. gam}marus are white or white\hl{-tipped . }

 The underside of the\hl{ claw of H}. americanus is orange or red , while that of H\hl{. gam}marus is creamy white or very pale red .

 = = Life cycle = =

 Female\hl{ H}. gammarus reach sexual maturity when they have grown to a carapace length of\hl{ 80 -- 85 millimet}res ( 3.1 -- 3.3 in ) , whereas males mature at a slightly smaller size\hl{ . Mating} typically occurs in summer between a recently moulted female , whose shell is therefore soft ,\hl{ and a} hard-shelled male . The\hl{ female carries} the eggs for up to 12 months , depending on the\hl{ temperature , attached to} her pleopods . Females carrying eggs are said to be " berried " and can\hl{ be found throughout} the year .

 The eggs hatch at night , and the larvae swim to the\hl{ water} surface where they drift with\hl{ the ocean} currents , preying on zooplankton . This stage involves three m\hl{oults and} lasts\hl{ for 15 --} 35 days . After the third moult , the juvenile takes on a form closer to the adult , and adopts\hl{ a} benthic lifestyle\hl{ . The juveniles} are rarely seen in the\hl{ wild , and} are poorly known , although they are known to be capable of digging extensive burrows . It is estimated that only 1 larva in every 20,000 survives to the benthic phase . When they reach a carapace length of 15\hl{ mm} ( 0.59 in ) , the juveniles leave their burrows and start their adult lives .

 = = Distribution = =

 Homarus gammarus is found across the north-eastern\hl{ Atlantic Ocean} from northern Norway to the Azores and Morocco , not including\hl{ the Baltic} Sea . It is also\hl{ present} in most of the Mediterranean Sea , only missing from the section east of Crete , and along only the\hl{ north}-west coast of\hl{ the} Black Sea . The northernmost populations are found in the Norwegian fjords Tysf
\end{tcolorbox}

\begin{tcolorbox}[breakable,title={$\sigma$-GPT-temporal}]
\footnotesize  the sea lobster , Eurybia

erythrocephala , which is also known as the water lobster , is found in the Caribbean Sea , southern Africa , the Ocean , Mediterranean Sea , the Azores , and the Mediterranean Sea . It is a small lobster , averaging 1 cm ( 0.4 in length of 1.in ) with a mass of 6 cm ( 2.4 in ) conspicuous pair of claws . The species ' main scavensters are crustaceans , with some individuals found in a variety of habitats .

@ dwelling in nature , the sea lobster is primarily a scavenger and is opportunistic , feeding plankton on crustaceanarus gammarus is a member of the group . It prefers areas with a large variety of water , such as rocky habitats , a variety of water bodies , shallow areas , and sandy shores , while the larger crustan lobster prefers muddy areas .

@ feeding is a seasonal activity , and is seasonal , with activity from November to March . It feeds on polychaetes and crustace in a wide variety of habitats , from sea to ocean -- 15 in the Atlantic Ocean , where it may reach 7 -- 12.5 cm ( 1.5 -- 3.9 in ) . Like other crustaceans , it has no claws and exoskeleton . Juveniles have a similar body shape to adults .

@ bodied , moulting occurs throughout the year . The juvenile stages are different from oysters , but hatch within a week . In the Mediterranean , it is a dielike species , with a length of between 5 and 15 cm ( 2.0 and . The maximum lifespan is around 45 years .

 mix of individual plankton and plankton species ; the sea lobster is vivip , which has two functional ovaries and two functional uteruses . Females produce one or two pups every other year .

@ sized populations of sea lobsters in the North Atlantic Oceanoskeleton is generally found in the southern Atlantic Ocean and Atlantic Ocean . One population in the British Isles has been found to have a female fetus . This occurs in a nest built on the sea floor , and the male then lays a clutch of several eggs . The eggs are laid up by 19 to 25.5 mm ( 0.75 relative of H. californica ) in size , with a plankton . The young measure 14 -- 16 mm ( 0.5 -- 0.6 in ) long . Males ranges do not appear to be capable of reproducing earlier in a number of stages , and the sex of a male sea lobster is variable . Reproduction is initiated by a single female , and in the spring or H. gammarica .

ocystidia are present in the skull of americanus . The first egg-laying is oval , measuring H. fusiformis by 10 -- 12 @@ tipped .

@ 8 , the claw of the second egg , is the largest of any lobster species ; H. californica may be larger than H. eurybia and the largest of the genus , while Female specimens are larger than the largest . The egg capsules measure 16 -- 20 mm ( 0 @ of 80 -- 85 millimetres ( 3.1 -- 3.1 in ) in length , with an average life body size . Moulting occurs at any time of year . The body is covered in mucous dermis , and the two pairs of gill slits . The female has an external vagina , which is in the center of the temperature , attached to the external vagina by a pair of ducts .

@ bearing spines that can be found near the mouth , from where they attach to the external gills . The eggs are the egg-shaped with the external gills , and are the only spines in the family that are fully moults and lasts for 15 -- 20 days . The female lays two to three pups in the spring , which are born in the summer afterts , returning to the lifestyle . The juveniles are relatively solitary in the wild , but some are found in groups .

 lines between the ages of 7 -- 18 years are known . The age of the oldest egg is usually around 22 -- 24 years . Females bear litters of 1 -- 4 eggs , which range from 8 -- 15 to 16 . The larvae are carnivorous , feeding on crustaceans , molluscs , and small crustaceans . In the Atlantic , their diet is dominated by crustaceans such as plankton in the eastern Atlantic Ocean , and crustaceans and crustaceans elsewhere , including the Mediterranean Sea . They may also prey on the shrimp Leucas selene , which is caught in the Mediterranean Sea . Off Newfoundland , the north Atlantic sea lobster is one of the many species that can be found in the North Atlantic , including the milk pike
\end{tcolorbox}

\begin{tcolorbox}[breakable,title={AC-GPT (ours)}]
\footnotesize  = = = Diet = = =

istic = = = =

Score = = = =  lobster , is a predator of the shrimp Tyroleanthys Ocean , Mediterranean herbarium , and Phyllocoryne shrimps . The exceptionally large carnivorous and very fast-swimming species length of H. gammar in general has a mass of 6 tonnes ( 7.6 tons ) conspicuous pair of claws . H. soca gamsters are smaller and lighter than H. capensis and H. tigrinus . Their canines grow up to 2.4 m ( 7.5 ft plankton ) , while Pharus gammari is a more stout and velvety lobster . H. soca is herbivorous and uses its long jaws to shelter from predators . This is an example of a species that uses a copan to make a burrow .

istic characteristics of H. soca include a small , rounded snout with a dorsal ridge ( caudal fin ) and a notch in the midline . The snout ends at 30 -- 15 in ( 7.5 -- 9 7 cm )-and @- 1.5 -- 3.5 in ) . Like other crustaceans , H. soca has exoskeleton that is covered by a layer of fine , thin , flexible armor . These features allow moulting and serve as a warning signal to predators and other crustacesters , but they also have a small countercurrent exchange of protection , allowing them to escape from predators . Males of H. soca are relatively slow , being more active than females . The head is broad and flattened with a flattened center . The dorsal ridge is also flexible and flexible ; the body is long and thin enough , which has a rounded tip .

450-3-3-1-0 structural architecture gives the species its name . The first phylloskeleton is thin , and when present it is in two layers : the outer layer is the exoskeleton and the outer layer is tape . This occurs in a few species of H. soca , notably Phyllocoryne shrimps , which can be made up by osmoregulation . The exoskeleton is also strongly modified by relative of H. soca 's , in contrast to the internal wall surface . The exoskeleton is thick and flexible and has a soft texture . H. soca has a few adaptations , which ranges do not occur in other species . The tail is as a number of different structures , including the tail which is roughly the length of the exoskeleton .

ufinaries are very different from the H. gammar species .

 cavity wall surfaces are the same colour as the americanus , but more egg-shaped . The shell of H. soca is flattened and oblong-tipped .

 stunned , with the claw of the male biting into the surface of the female , is called H. soca mucilidum . Males develop two pairs of long spines on the underside of each . Female pinnacles attach to the spines and the female pinnacles attach to the underside of 80 -- 85 millimetres ( 3.1 -- 3.3 in ) spines on the underside of her own size . Moulting takes place over two days . The female pinnacles attach to the spines , and can be up to 1 metre ( 3 ft The female also can attach to the spines at any time of the temperature , attached to the spines , and on the underside of her legs .

 slamings are can be found on the underside of the pinnacles . When feeding , females move the eggs along the bottom of the sea , with the eggs excavated in the water . The eggs hatch after about 25 days . The moults typically lasts for 15 -- 20 days and are incubated for about 100 days .

 VP

 gramme or gramme ists are unusual in their lifestyle . The eggs hatch in water in the wild , but they are not protected by the Ministry of Marine Defence ( MINAE ) . The scop shell is sometimes used for storing the eggs .

 suffered from a succession of problems , mostly due to loss of fertility . Two hatchlings were caught and 15 -- 20 \% of the egg was lost to the sea . The second hatchling , an egg-laying worker , died after 20 days .

 1440-1450 H. soca in the eastern Atlantic was caught in January 1985 and cared for for many years , including the installation of a nest and also spent time in shallow water .

 Calling lobster ( Chionella europaea ) is found in the north Atlantic Ocean . The lifespan of H. soca is around 15 to 20 years .

 punched-up
\end{tcolorbox}

\begin{tcolorbox}[breakable,title={$\sigma$-GPT-random}]
\footnotesize  = = = Description = =

 Cactidar , which means

 of " jelly , lobster , is a type-dwelling to in the Atlantic Ocean , Mediterranean Sea , and a family of many other fishes . It is a largeed body , a slender snout and a body , with a length of about 3 centimeters in ) and a mass of 6 in ( 13 in ) . Its head and conspicuous pair of claws , and its longwater lobsters are small-developed with , flat-like extensions . The mouth is a grooved-like . stalk that is not very large , and in the plankton and the genus Somarus gamelai is a memberel of of therop order " claw " " for the c g "halace the @ods the like with c atenoph like the the c c P the. membranous and and the the c = = = Description  Adult T " = = = c The c = b a Squ H c  3 7 c c g C ( Found in " of rows 4 L The 7 Squ : 11 -- 15 in ) wide 5 T 2 lob g : 7 -- 6-5ans -- 1.5 -- 3 2 5 Sp 16 3 3 . Like other crustaceans , 5 7 C P The exoskeleton @ 7 7 Sp 9 T 12 @ 4 = = 10 g 1 P b 1 moulton  4 c M 1 C S c c Small freshwater oysters , but other g = c T C T 7 b Squ Unicorn Unicorn g Th Ch c Sp B c C  Par " 5 P c = c m . Sp " 3 . The p Wal Squ Squ D P Sp Sc C Sc C T Sc Squ Hydro Sp Sh " ;

 C Som T T 2 , which has a

 Squ Hyd = = Sh Squ Th Sp S Nem? Squ @ S Sp c Sp Squ C Wal S Squ C Squ = P O Th C C Nem Nemoskeleton isps

 T Cl Sp Squ Sc C C Squ C c Squ Squ Unicorn T T T Th Sp Wal The D Sty 4 . This occurs in B On P P S T T Unicorn Squ P T Hyd Sc Sty C C L Hyd T  Ar A Found up by a , P Unicorn P B Sp Squ Nem Squ T Sp Small The relative of Heteridae Squ B D Black S " T @ Th Sty 4 . The  S Th Unicorn Squ T T Sty Squ P T T C Sc M Squ Squ S the  The T with ranges do not C Squ Sc S Squ C Most? On a number  In Like The In P Sp Ar the

 In P The the

 Unicorn Black T Sp Sc T Black

 or H. gammarina =

 Unicorn Se P the a  S the the americanus , The The 3-foot ch  = , H.  T A H @ the Water-tipped .

 The Order of the claw of the the.   the M  a = = The H.  =- ,- @  =? = @ @a = @ = of of

 = Female  the H = = @us.

 , =

 . =

 =let order of 80 -- 85 mill @ = = = @ ( ( = Sc = = = =a us a the = = @ in Is size . M

 = = Unicorn ' S S @ = ,  , T Sp. M = Unicorn , and of  = , ,a =

 = The female @  , S P = . The = Because In the temperature , attached to = = , ofed ,- = @ = and =ed = T Or Unicorn can be found  Adult = Hyd Unicorn =

 = = = Unicorn = = = Cidae of the ' the = Dip As with the @ The T Dip =  the = the = =-  Most The Adult moults , lasts for 15 years  Sea = the S Hyd = = = of

 of Hyd the @ = S P @ = sea Ots = In In in lifestyle . The

 Hyd In T In the wild , D C Hydro Lep = =  S Sp . C @ C S idae and Tes = Hyd Middle  = @ Squ Sc = C? Sp C T Hyd Nem P c  C  Squ In = = T  P = Sea

 15 Ch Unicorn = C T The Most  Squ Squ Dip T? Dip T = D Sp c  Middle Sp

 The  C In  = L in T Squ  =   Most = , " North of the eastern Atlantic @   Middle Squ Lep H Sp ( Middle Middle In including the   In S Unicorn also and  the Sp  Sp Squ Sc L T . =n let  = New = " In the north g Or T Unicorn = of  =     Sc Unicorn of = ,  of  of Hydroa
\end{tcolorbox}

%% file: ablation_table.tex
\begin{table*}[ht]
\caption{Ablation Study (Small): Perplexity ($\downarrow$) across evaluation modes, model types, and conditioning scales using GPT-2 Small (82M) backbone.}
\label{tab:ablation_small}
\begin{center}
\resizebox{\columnwidth}{!}{%
\begin{tabular}{ll ccccc}
\toprule
\textbf{Scale} & \textbf{Model} & \textbf{Unconditional} & \textbf{Infilling} & \textbf{Training dist.} & \textbf{Infilling w/o future} & \textbf{Training dist. w/o future} \\
\midrule
\multirow{5}{*}{cond0-20}
  & AC-GPT (ours) & 18.0 {\scriptsize $\pm$ 0.0} & \textbf{17.2 {\scriptsize $\pm$ 0.1}} & \textbf{16.5 {\scriptsize $\pm$ 0.1}} & \textbf{17.5 {\scriptsize $\pm$ 0.1}} & 17.9 {\scriptsize $\pm$ 0.1} \\
  & $\sigma$-GPT-temporal & 18.1 {\scriptsize $\pm$ 0.0} & 17.4 {\scriptsize $\pm$ 0.0} & 17.0 {\scriptsize $\pm$ 0.1} & 17.7 {\scriptsize $\pm$ 0.1} & 18.1 {\scriptsize $\pm$ 0.1} \\
  & $\sigma$-GPT-random & 194.8 {\scriptsize $\pm$ 94.7} & 199.6 {\scriptsize $\pm$ 119.1} & 57.8 {\scriptsize $\pm$ 3.3} & 179.2 {\scriptsize $\pm$ 85.1} & 192.1 {\scriptsize $\pm$ 93.7} \\
  & MLM (BERT) & 62.3 {\scriptsize $\pm$ 6.3} & 60.0 {\scriptsize $\pm$ 6.0} & 56.2 {\scriptsize $\pm$ 6.0} & 60.9 {\scriptsize $\pm$ 6.0} & 61.8 {\scriptsize $\pm$ 6.2} \\
  & GPT & \textbf{17.9 {\scriptsize $\pm$ 0.0}} & -- & -- & 17.5 {\scriptsize $\pm$ 0.1} & \textbf{17.9 {\scriptsize $\pm$ 0.0}} \\
\cmidrule{1-7}
\multirow{5}{*}{cond0-40}
  & AC-GPT (ours) & 18.2 {\scriptsize $\pm$ 0.0} & \textbf{17.2 {\scriptsize $\pm$ 0.1}} & \textbf{15.8 {\scriptsize $\pm$ 0.1}} & 17.6 {\scriptsize $\pm$ 0.1} & 18.1 {\scriptsize $\pm$ 0.1} \\
  & $\sigma$-GPT-temporal & 18.5 {\scriptsize $\pm$ 0.1} & 17.3 {\scriptsize $\pm$ 0.1} & 16.4 {\scriptsize $\pm$ 0.1} & 17.9 {\scriptsize $\pm$ 0.2} & 18.4 {\scriptsize $\pm$ 0.1} \\
  & $\sigma$-GPT-random & 231.9 {\scriptsize $\pm$ 196.1} & 323.4 {\scriptsize $\pm$ 318.0} & 55.8 {\scriptsize $\pm$ 2.7} & 226.0 {\scriptsize $\pm$ 199.2} & 230.5 {\scriptsize $\pm$ 197.5} \\
  & MLM (BERT) & 52.7 {\scriptsize $\pm$ 1.1} & 50.0 {\scriptsize $\pm$ 1.1} & 44.9 {\scriptsize $\pm$ 0.9} & 51.3 {\scriptsize $\pm$ 1.2} & 52.3 {\scriptsize $\pm$ 0.8} \\
  & GPT & \textbf{17.9 {\scriptsize $\pm$ 0.0}} & -- & -- & \textbf{17.4 {\scriptsize $\pm$ 0.1}} & \textbf{17.8 {\scriptsize $\pm$ 0.0}} \\
\cmidrule{1-7}
\multirow{5}{*}{cond0-60}
  & AC-GPT (ours) & 18.4 {\scriptsize $\pm$ 0.0} & \textbf{17.1 {\scriptsize $\pm$ 0.2}} & \textbf{15.2 {\scriptsize $\pm$ 0.2}} & 17.6 {\scriptsize $\pm$ 0.2} & 18.1 {\scriptsize $\pm$ 0.1} \\
  & $\sigma$-GPT-temporal & 18.8 {\scriptsize $\pm$ 0.1} & 17.3 {\scriptsize $\pm$ 0.1} & 15.9 {\scriptsize $\pm$ 0.2} & 18.1 {\scriptsize $\pm$ 0.1} & 18.6 {\scriptsize $\pm$ 0.2} \\
  & $\sigma$-GPT-random & 234.0 {\scriptsize $\pm$ 82.5} & 315.1 {\scriptsize $\pm$ 169.3} & 55.2 {\scriptsize $\pm$ 2.3} & 237.6 {\scriptsize $\pm$ 109.1} & 233.1 {\scriptsize $\pm$ 86.4} \\
  & MLM (BERT) & 48.4 {\scriptsize $\pm$ 1.1} & 45.4 {\scriptsize $\pm$ 0.8} & 39.2 {\scriptsize $\pm$ 1.5} & 46.9 {\scriptsize $\pm$ 0.8} & 47.7 {\scriptsize $\pm$ 1.4} \\
  & GPT & \textbf{17.9 {\scriptsize $\pm$ 0.0}} & -- & -- & \textbf{17.4 {\scriptsize $\pm$ 0.1}} & \textbf{17.7 {\scriptsize $\pm$ 0.1}} \\
\cmidrule{1-7}
\multirow{5}{*}{cond0-80}
  & AC-GPT (ours) & 18.6 {\scriptsize $\pm$ 0.1} & \textbf{17.2 {\scriptsize $\pm$ 0.1}} & \textbf{14.9 {\scriptsize $\pm$ 0.2}} & 17.9 {\scriptsize $\pm$ 0.0} & 18.3 {\scriptsize $\pm$ 0.3} \\
  & $\sigma$-GPT-temporal & 19.1 {\scriptsize $\pm$ 0.1} & 17.4 {\scriptsize $\pm$ 0.1} & 15.6 {\scriptsize $\pm$ 0.3} & 18.5 {\scriptsize $\pm$ 0.1} & 19.0 {\scriptsize $\pm$ 0.4} \\
  & $\sigma$-GPT-random & 178.6 {\scriptsize $\pm$ 19.8} & 240.9 {\scriptsize $\pm$ 27.6} & 53.8 {\scriptsize $\pm$ 1.2} & 157.2 {\scriptsize $\pm$ 17.1} & 175.5 {\scriptsize $\pm$ 21.6} \\
  & MLM (BERT) & 45.3 {\scriptsize $\pm$ 1.4} & 42.0 {\scriptsize $\pm$ 1.4} & 35.5 {\scriptsize $\pm$ 1.6} & 43.9 {\scriptsize $\pm$ 1.3} & 44.8 {\scriptsize $\pm$ 1.9} \\
  & GPT & \textbf{17.9 {\scriptsize $\pm$ 0.0}} & -- & -- & \textbf{17.5 {\scriptsize $\pm$ 0.0}} & \textbf{17.8 {\scriptsize $\pm$ 0.4}} \\
\cmidrule{1-7}
\multirow{5}{*}{cond0-100}
  & AC-GPT (ours) & 18.8 {\scriptsize $\pm$ 0.1} & \textbf{17.4 {\scriptsize $\pm$ 0.1}} & \textbf{14.7 {\scriptsize $\pm$ 0.1}} & 18.1 {\scriptsize $\pm$ 0.1} & 18.6 {\scriptsize $\pm$ 0.2} \\
  & $\sigma$-GPT-temporal & 19.6 {\scriptsize $\pm$ 0.1} & 17.7 {\scriptsize $\pm$ 0.1} & 15.6 {\scriptsize $\pm$ 0.1} & 19.0 {\scriptsize $\pm$ 0.2} & 19.4 {\scriptsize $\pm$ 0.3} \\
  & $\sigma$-GPT-random & 188.5 {\scriptsize $\pm$ 14.2} & 193.3 {\scriptsize $\pm$ 15.9} & 54.2 {\scriptsize $\pm$ 0.6} & 151.9 {\scriptsize $\pm$ 28.4} & 180.2 {\scriptsize $\pm$ 18.2} \\
  & MLM (BERT) & 45.4 {\scriptsize $\pm$ 0.7} & 41.8 {\scriptsize $\pm$ 0.7} & 34.7 {\scriptsize $\pm$ 0.4} & 43.9 {\scriptsize $\pm$ 0.7} & 44.8 {\scriptsize $\pm$ 1.0} \\
  & GPT & \textbf{17.9 {\scriptsize $\pm$ 0.0}} & -- & -- & \textbf{17.5 {\scriptsize $\pm$ 0.2}} & \textbf{17.8 {\scriptsize $\pm$ 0.2}} \\
\bottomrule
\end{tabular}%
}
\end{center}
\end{table*}

\begin{table*}[ht]
\caption{Ablation Study (Base): Perplexity ($\downarrow$) across evaluation modes, model types, and conditioning scales using GPT-2 Base (124M) backbone.}
\label{tab:ablation_base}
\begin{center}
\resizebox{\columnwidth}{!}{%
\begin{tabular}{ll ccccc}
\toprule
\textbf{Scale} & \textbf{Model} & \textbf{Unconditional} & \textbf{Infilling} & \textbf{Training dist.} & \textbf{Infilling w/o future} & \textbf{Training dist. w/o future} \\
\midrule
\multirow{5}{*}{cond0-20}
  & AC-GPT (ours) & 16.7 {\scriptsize $\pm$ 0.0} & 15.9 {\scriptsize $\pm$ 0.1} & \textbf{15.2 {\scriptsize $\pm$ 0.1}} & 16.2 {\scriptsize $\pm$ 0.1} & 16.6 {\scriptsize $\pm$ 0.1} \\
  & $\sigma$-GPT-temporal & \textbf{16.6 {\scriptsize $\pm$ 0.0}} & \textbf{15.9 {\scriptsize $\pm$ 0.0}} & 15.5 {\scriptsize $\pm$ 0.0} & \textbf{16.2 {\scriptsize $\pm$ 0.1}} & \textbf{16.6 {\scriptsize $\pm$ 0.0}} \\
  & $\sigma$-GPT-random & 405.0 {\scriptsize $\pm$ 17.8} & 656.3 {\scriptsize $\pm$ 65.1} & 39.1 {\scriptsize $\pm$ 0.8} & 404.0 {\scriptsize $\pm$ 21.1} & 403.6 {\scriptsize $\pm$ 17.8} \\
  & MLM (BERT) & 49.8 {\scriptsize $\pm$ 0.8} & 47.7 {\scriptsize $\pm$ 0.9} & 44.5 {\scriptsize $\pm$ 0.7} & 48.7 {\scriptsize $\pm$ 0.8} & 49.4 {\scriptsize $\pm$ 0.8} \\
  & GPT & 16.7 {\scriptsize $\pm$ 0.1} & -- & -- & 16.3 {\scriptsize $\pm$ 0.1} & 16.6 {\scriptsize $\pm$ 0.0} \\
\cmidrule{1-7}
\multirow{5}{*}{cond0-40}
  & AC-GPT (ours) & 16.7 {\scriptsize $\pm$ 0.0} & \textbf{15.7 {\scriptsize $\pm$ 0.1}} & \textbf{14.4 {\scriptsize $\pm$ 0.1}} & \textbf{16.1 {\scriptsize $\pm$ 0.1}} & \textbf{16.6 {\scriptsize $\pm$ 0.1}} \\
  & $\sigma$-GPT-temporal & 16.9 {\scriptsize $\pm$ 0.1} & 15.8 {\scriptsize $\pm$ 0.1} & 14.8 {\scriptsize $\pm$ 0.1} & 16.4 {\scriptsize $\pm$ 0.2} & 16.8 {\scriptsize $\pm$ 0.1} \\
  & $\sigma$-GPT-random & 376.2 {\scriptsize $\pm$ 52.9} & 814.7 {\scriptsize $\pm$ 321.5} & 35.9 {\scriptsize $\pm$ 1.1} & 390.7 {\scriptsize $\pm$ 63.1} & 377.5 {\scriptsize $\pm$ 54.6} \\
  & MLM (BERT) & 42.6 {\scriptsize $\pm$ 0.5} & 40.3 {\scriptsize $\pm$ 0.7} & 35.9 {\scriptsize $\pm$ 0.3} & 41.4 {\scriptsize $\pm$ 0.6} & 42.3 {\scriptsize $\pm$ 0.3} \\
  & GPT & \textbf{16.7 {\scriptsize $\pm$ 0.1}} & -- & -- & 16.2 {\scriptsize $\pm$ 0.2} & 16.6 {\scriptsize $\pm$ 0.0} \\
\cmidrule{1-7}
\multirow{5}{*}{cond0-60}
  & AC-GPT (ours) & 16.8 {\scriptsize $\pm$ 0.0} & \textbf{15.6 {\scriptsize $\pm$ 0.2}} & \textbf{13.8 {\scriptsize $\pm$ 0.2}} & \textbf{16.1 {\scriptsize $\pm$ 0.2}} & 16.6 {\scriptsize $\pm$ 0.1} \\
  & $\sigma$-GPT-temporal & 17.2 {\scriptsize $\pm$ 0.1} & 15.7 {\scriptsize $\pm$ 0.2} & 14.3 {\scriptsize $\pm$ 0.2} & 16.6 {\scriptsize $\pm$ 0.3} & 17.0 {\scriptsize $\pm$ 0.0} \\
  & $\sigma$-GPT-random & 487.6 {\scriptsize $\pm$ 119.0} & 940.5 {\scriptsize $\pm$ 438.1} & 35.2 {\scriptsize $\pm$ 1.9} & 558.1 {\scriptsize $\pm$ 175.3} & 493.4 {\scriptsize $\pm$ 120.9} \\
  & MLM (BERT) & 39.1 {\scriptsize $\pm$ 0.3} & 36.3 {\scriptsize $\pm$ 0.6} & 31.1 {\scriptsize $\pm$ 0.3} & 37.7 {\scriptsize $\pm$ 0.6} & 38.4 {\scriptsize $\pm$ 0.0} \\
  & GPT & \textbf{16.7 {\scriptsize $\pm$ 0.1}} & -- & -- & 16.2 {\scriptsize $\pm$ 0.1} & \textbf{16.5 {\scriptsize $\pm$ 0.2}} \\
\cmidrule{1-7}
\multirow{5}{*}{cond0-80}
  & AC-GPT (ours) & 17.0 {\scriptsize $\pm$ 0.1} & \textbf{15.7 {\scriptsize $\pm$ 0.1}} & \textbf{13.5 {\scriptsize $\pm$ 0.1}} & 16.3 {\scriptsize $\pm$ 0.1} & 16.7 {\scriptsize $\pm$ 0.3} \\
  & $\sigma$-GPT-temporal & 17.4 {\scriptsize $\pm$ 0.1} & 15.8 {\scriptsize $\pm$ 0.0} & 14.0 {\scriptsize $\pm$ 0.2} & 16.8 {\scriptsize $\pm$ 0.0} & 17.2 {\scriptsize $\pm$ 0.4} \\
  & $\sigma$-GPT-random & 414.2 {\scriptsize $\pm$ 310.0} & 580.7 {\scriptsize $\pm$ 438.5} & 34.7 {\scriptsize $\pm$ 2.9} & 486.9 {\scriptsize $\pm$ 367.9} & 430.0 {\scriptsize $\pm$ 325.8} \\
  & MLM (BERT) & 37.9 {\scriptsize $\pm$ 0.4} & 35.0 {\scriptsize $\pm$ 0.4} & 29.2 {\scriptsize $\pm$ 0.6} & 36.7 {\scriptsize $\pm$ 0.4} & 37.4 {\scriptsize $\pm$ 0.9} \\
  & GPT & \textbf{16.7 {\scriptsize $\pm$ 0.1}} & -- & -- & \textbf{16.3 {\scriptsize $\pm$ 0.1}} & \textbf{16.5 {\scriptsize $\pm$ 0.4}} \\
\cmidrule{1-7}
\multirow{5}{*}{cond0-100}
  & AC-GPT (ours) & 17.3 {\scriptsize $\pm$ 0.2} & \textbf{15.9 {\scriptsize $\pm$ 0.1}} & \textbf{13.4 {\scriptsize $\pm$ 0.1}} & 16.6 {\scriptsize $\pm$ 0.0} & 17.1 {\scriptsize $\pm$ 0.2} \\
  & $\sigma$-GPT-temporal & 17.6 {\scriptsize $\pm$ 0.1} & 15.9 {\scriptsize $\pm$ 0.2} & 13.8 {\scriptsize $\pm$ 0.1} & 17.1 {\scriptsize $\pm$ 0.3} & 17.5 {\scriptsize $\pm$ 0.3} \\
  & $\sigma$-GPT-random & 415.0 {\scriptsize $\pm$ 110.0} & 600.0 {\scriptsize $\pm$ 42.1} & 34.2 {\scriptsize $\pm$ 3.2} & 459.7 {\scriptsize $\pm$ 204.8} & 418.8 {\scriptsize $\pm$ 135.7} \\
  & MLM (BERT) & 38.1 {\scriptsize $\pm$ 0.1} & 34.8 {\scriptsize $\pm$ 0.3} & 28.5 {\scriptsize $\pm$ 0.3} & 36.8 {\scriptsize $\pm$ 0.3} & 37.6 {\scriptsize $\pm$ 0.5} \\
  & GPT & \textbf{16.7 {\scriptsize $\pm$ 0.1}} & -- & -- & \textbf{16.3 {\scriptsize $\pm$ 0.2}} & \textbf{16.6 {\scriptsize $\pm$ 0.2}} \\
\bottomrule
\end{tabular}%
}
\end{center}
\end{table*}

\begin{table*}[ht]
\caption{Ablation Study (Medium): Perplexity ($\downarrow$) across evaluation modes, model types, and conditioning scales using GPT-2 Medium (204M) backbone.}
\label{tab:ablation_medium}
\begin{center}
\resizebox{\columnwidth}{!}{%
\begin{tabular}{ll ccccc}
\toprule
\textbf{Scale} & \textbf{Model} & \textbf{Unconditional} & \textbf{Infilling} & \textbf{Training dist.} & \textbf{Infilling w/o future} & \textbf{Training dist. w/o future} \\
\midrule
\multirow{5}{*}{cond0-20}
  & AC-GPT (ours) & 16.9 {\scriptsize $\pm$ 0.1} & 16.1 {\scriptsize $\pm$ 0.1} & 15.4 {\scriptsize $\pm$ 0.1} & 16.4 {\scriptsize $\pm$ 0.1} & 16.8 {\scriptsize $\pm$ 0.1} \\
  & $\sigma$-GPT-temporal & \textbf{16.5 {\scriptsize $\pm$ 0.1}} & \textbf{15.9 {\scriptsize $\pm$ 0.1}} & \textbf{15.4 {\scriptsize $\pm$ 0.0}} & \textbf{16.1 {\scriptsize $\pm$ 0.1}} & \textbf{16.5 {\scriptsize $\pm$ 0.1}} \\
  & $\sigma$-GPT-random & 449.7 {\scriptsize $\pm$ 241.9} & 674.8 {\scriptsize $\pm$ 196.8} & 32.4 {\scriptsize $\pm$ 1.1} & 479.4 {\scriptsize $\pm$ 275.9} & 454.7 {\scriptsize $\pm$ 248.3} \\
  & MLM (BERT) & 44.1 {\scriptsize $\pm$ 0.4} & 42.2 {\scriptsize $\pm$ 0.3} & 39.3 {\scriptsize $\pm$ 0.2} & 42.7 {\scriptsize $\pm$ 0.2} & 43.7 {\scriptsize $\pm$ 0.3} \\
  & GPT & 17.2 {\scriptsize $\pm$ 0.0} & -- & -- & 16.7 {\scriptsize $\pm$ 0.0} & 17.1 {\scriptsize $\pm$ 0.0} \\
\cmidrule{1-7}
\multirow{5}{*}{cond0-40}
  & AC-GPT (ours) & 16.7 {\scriptsize $\pm$ 0.0} & 15.7 {\scriptsize $\pm$ 0.2} & \textbf{14.4 {\scriptsize $\pm$ 0.1}} & 16.1 {\scriptsize $\pm$ 0.2} & 16.5 {\scriptsize $\pm$ 0.1} \\
  & $\sigma$-GPT-temporal & \textbf{16.4 {\scriptsize $\pm$ 0.1}} & \textbf{15.5 {\scriptsize $\pm$ 0.1}} & 14.5 {\scriptsize $\pm$ 0.1} & \textbf{15.9 {\scriptsize $\pm$ 0.1}} & \textbf{16.3 {\scriptsize $\pm$ 0.2}} \\
  & $\sigma$-GPT-random & 579.9 {\scriptsize $\pm$ 140.7} & 835.5 {\scriptsize $\pm$ 29.0} & 30.2 {\scriptsize $\pm$ 0.4} & 658.3 {\scriptsize $\pm$ 159.6} & 590.9 {\scriptsize $\pm$ 147.8} \\
  & MLM (BERT) & 37.6 {\scriptsize $\pm$ 0.3} & 35.4 {\scriptsize $\pm$ 0.5} & 31.4 {\scriptsize $\pm$ 0.5} & 36.4 {\scriptsize $\pm$ 0.5} & 37.2 {\scriptsize $\pm$ 0.3} \\
  & GPT & 17.2 {\scriptsize $\pm$ 0.0} & -- & -- & 16.7 {\scriptsize $\pm$ 0.2} & 17.1 {\scriptsize $\pm$ 0.1} \\
\cmidrule{1-7}
\multirow{5}{*}{cond0-60}
  & AC-GPT (ours) & 16.6 {\scriptsize $\pm$ 0.1} & 15.4 {\scriptsize $\pm$ 0.2} & \textbf{13.5 {\scriptsize $\pm$ 0.2}} & 15.9 {\scriptsize $\pm$ 0.2} & 16.3 {\scriptsize $\pm$ 0.1} \\
  & $\sigma$-GPT-temporal & \textbf{16.4 {\scriptsize $\pm$ 0.1}} & \textbf{15.2 {\scriptsize $\pm$ 0.2}} & 13.7 {\scriptsize $\pm$ 0.3} & \textbf{15.9 {\scriptsize $\pm$ 0.1}} & \textbf{16.2 {\scriptsize $\pm$ 0.2}} \\
  & $\sigma$-GPT-random & 532.2 {\scriptsize $\pm$ 397.7} & 764.2 {\scriptsize $\pm$ 573.1} & 28.5 {\scriptsize $\pm$ 0.4} & 661.4 {\scriptsize $\pm$ 517.3} & 545.4 {\scriptsize $\pm$ 410.4} \\
  & MLM (BERT) & 35.4 {\scriptsize $\pm$ 0.8} & 32.8 {\scriptsize $\pm$ 0.7} & 27.9 {\scriptsize $\pm$ 1.1} & 34.1 {\scriptsize $\pm$ 0.7} & 34.8 {\scriptsize $\pm$ 1.0} \\
  & GPT & 17.2 {\scriptsize $\pm$ 0.0} & -- & -- & 16.6 {\scriptsize $\pm$ 0.2} & 17.0 {\scriptsize $\pm$ 0.1} \\
\cmidrule{1-7}
\multirow{5}{*}{cond0-80}
  & AC-GPT (ours) & 16.5 {\scriptsize $\pm$ 0.0} & 15.3 {\scriptsize $\pm$ 0.1} & \textbf{13.1 {\scriptsize $\pm$ 0.2}} & \textbf{16.0 {\scriptsize $\pm$ 0.0}} & \textbf{16.3 {\scriptsize $\pm$ 0.3}} \\
  & $\sigma$-GPT-temporal & \textbf{16.5 {\scriptsize $\pm$ 0.1}} & \textbf{15.2 {\scriptsize $\pm$ 0.1}} & 13.4 {\scriptsize $\pm$ 0.2} & 16.0 {\scriptsize $\pm$ 0.1} & 16.4 {\scriptsize $\pm$ 0.4} \\
  & $\sigma$-GPT-random & 479.6 {\scriptsize $\pm$ 215.7} & 896.5 {\scriptsize $\pm$ 315.4} & 28.4 {\scriptsize $\pm$ 1.4} & 588.7 {\scriptsize $\pm$ 306.9} & 501.8 {\scriptsize $\pm$ 225.1} \\
  & MLM (BERT) & 34.2 {\scriptsize $\pm$ 0.9} & 31.6 {\scriptsize $\pm$ 0.8} & 26.2 {\scriptsize $\pm$ 1.1} & 33.1 {\scriptsize $\pm$ 0.9} & 33.8 {\scriptsize $\pm$ 1.2} \\
  & GPT & 17.2 {\scriptsize $\pm$ 0.0} & -- & -- & 16.7 {\scriptsize $\pm$ 0.1} & 17.0 {\scriptsize $\pm$ 0.4} \\
\cmidrule{1-7}
\multirow{5}{*}{cond0-100}
  & AC-GPT (ours) & \textbf{16.6 {\scriptsize $\pm$ 0.1}} & 15.3 {\scriptsize $\pm$ 0.2} & \textbf{12.7 {\scriptsize $\pm$ 0.1}} & \textbf{16.0 {\scriptsize $\pm$ 0.2}} & \textbf{16.3 {\scriptsize $\pm$ 0.2}} \\
  & $\sigma$-GPT-temporal & 16.6 {\scriptsize $\pm$ 0.0} & \textbf{15.2 {\scriptsize $\pm$ 0.2}} & 13.0 {\scriptsize $\pm$ 0.1} & 16.1 {\scriptsize $\pm$ 0.2} & 16.4 {\scriptsize $\pm$ 0.2} \\
  & $\sigma$-GPT-random & 451.8 {\scriptsize $\pm$ 163.4} & 881.6 {\scriptsize $\pm$ 151.3} & 29.0 {\scriptsize $\pm$ 1.2} & 540.4 {\scriptsize $\pm$ 240.7} & 461.6 {\scriptsize $\pm$ 170.1} \\
  & MLM (BERT) & 34.3 {\scriptsize $\pm$ 0.8} & 31.4 {\scriptsize $\pm$ 1.0} & 25.5 {\scriptsize $\pm$ 0.9} & 33.1 {\scriptsize $\pm$ 1.1} & 33.9 {\scriptsize $\pm$ 1.1} \\
  & GPT & 17.2 {\scriptsize $\pm$ 0.0} & -- & -- & 16.7 {\scriptsize $\pm$ 0.3} & 17.0 {\scriptsize $\pm$ 0.2} \\
\bottomrule
\end{tabular}%
}
\end{center}
\end{table*}